\documentclass[conference]{IEEEtran}
\IEEEoverridecommandlockouts
\usepackage{cite}
\usepackage{amsmath,amssymb,amsfonts}
\usepackage{algorithmic}
\usepackage{graphicx}
\usepackage{textcomp}
\usepackage{xcolor}
\def\BibTeX{{\rm B\kern-.05em{\sc i\kern-.025em b}\kern-.08em
    T\kern-.1667em\lower.7ex\hbox{E}\kern-.125emX}}
\usepackage{subfigure}
\usepackage{booktabs}

\usepackage[algo2e,ruled,linesnumbered]{algorithm2e}
\usepackage{multirow}
\usepackage{hyperref}
\newcommand{\ip}[2]{\left\langle #1, #2 \right \rangle}

\newtheorem{theorem}{Theorem}[section]
\newtheorem{lemma}{Lemma}[section]

\newtheorem{assumption}{Assumption}[section]

\SetKwFor{While}{while}{}{end while}%

\begin{document}

\title{Highly Scalable and Provably Accurate Classification in Poincar\'e Balls
}

\author{
\IEEEauthorblockN{Eli Chien\textsuperscript{*}, Chao Pan\textsuperscript{*}\thanks{*equal contribution}, Puoya Tabaghi, Olgica Milenkovic}
\IEEEauthorblockA{\textit{ECE, University of Illinois Urbana-Champaign, 306 N Wright St, Urbana, Illinois 61801, USA} \\
\{ichien3, chaopan2, tabaghi2, milenkov\}@illinois.edu}
}

\maketitle

\begin{abstract}
Many high-dimensional and large-volume data sets of practical relevance have hierarchical structures induced by trees, graphs or time series. Such data sets are hard to process in Euclidean spaces and one often seeks low-dimensional embeddings in other space forms to perform required learning tasks. For hierarchical data, the space of choice is a hyperbolic space since it guarantees low-distortion embeddings for tree-like structures. Unfortunately, the geometry of hyperbolic spaces has properties not encountered in Euclidean spaces that pose challenges when trying to rigorously analyze algorithmic solutions. Here, for the first time, we establish a unified framework for learning scalable and simple hyperbolic linear classifiers with provable performance guarantees. The gist of our approach is to focus on Poincar\'e ball models and formulate the classification problems using tangent space formalisms. Our results include a new hyperbolic and second-order perceptron algorithm as well as an efficient and highly accurate convex optimization setup for hyperbolic support vector machine classifiers. All algorithms provably converge and are highly scalable as they have complexities comparable to those of their Euclidean counterparts. Their performance accuracies on synthetic data sets comprising millions of points, as well as on complex real-world data sets such as single-cell RNA-seq expression measurements, CIFAR10, Fashion-MNIST and mini-ImageNet. Our code can be found at: \url{https://github.com/thupchnsky/PoincareLinearClassification}.~\footnote{A short version of this paper is accepted by ICDM 2021 as a regular paper.} 
\end{abstract}


\section{Introduction}
Representation learning in hyperbolic spaces has received significant interest due to its effectiveness in capturing latent hierarchical structures~\cite{krioukov2010hyperbolic,sarkar2011low,pmlr-v80-sala18a,nickel2017poincare,papadopoulos2015network,tifrea2018poincare}. It is known that arbitrarily low-distortion embeddings of tree-structures in Euclidean spaces is impossible even when using an unbounded number of dimensions~\cite{linial1995geometry}. In contrast, precise and simple embeddings are possible in the Poincar\'e disk, a hyperbolic space model with only two dimensions~\cite{sarkar2011low}. 

Despite their representational power, hyperbolic spaces are still lacking foundational analytical results and algorithmic solutions for a wide variety of downstream machine learning tasks. In particular, the question of designing highly-scalable classification algorithms with provable performance guarantees that exploit the structure of hyperbolic spaces remains open. While a few prior works have proposed specific algorithms for learning classifiers in hyperbolic space, they are primarily empirical in nature and do not come with theoretical convergence guarantees~\cite{cho2019large,monath2019gradient}. The work~\cite{weber2020robust} described the first attempt to establish performance guarantees for the hyperboloid perceptron, but the proposed algorithm is not transparent and fails to converge in practice. Furthermore, the methodology used does not naturally generalize to other important classification methods such as support vector machines (SVMs)~\cite{cortes1995support}. Hence, a natural question arises: Is there a unified framework that allows one to generalize classification algorithms for Euclidean spaces to hyperbolic spaces, make them highly scalable and rigorously establish their performance guarantees?

\begin{figure}[t]
  \center
  \includegraphics[width=1 \linewidth]{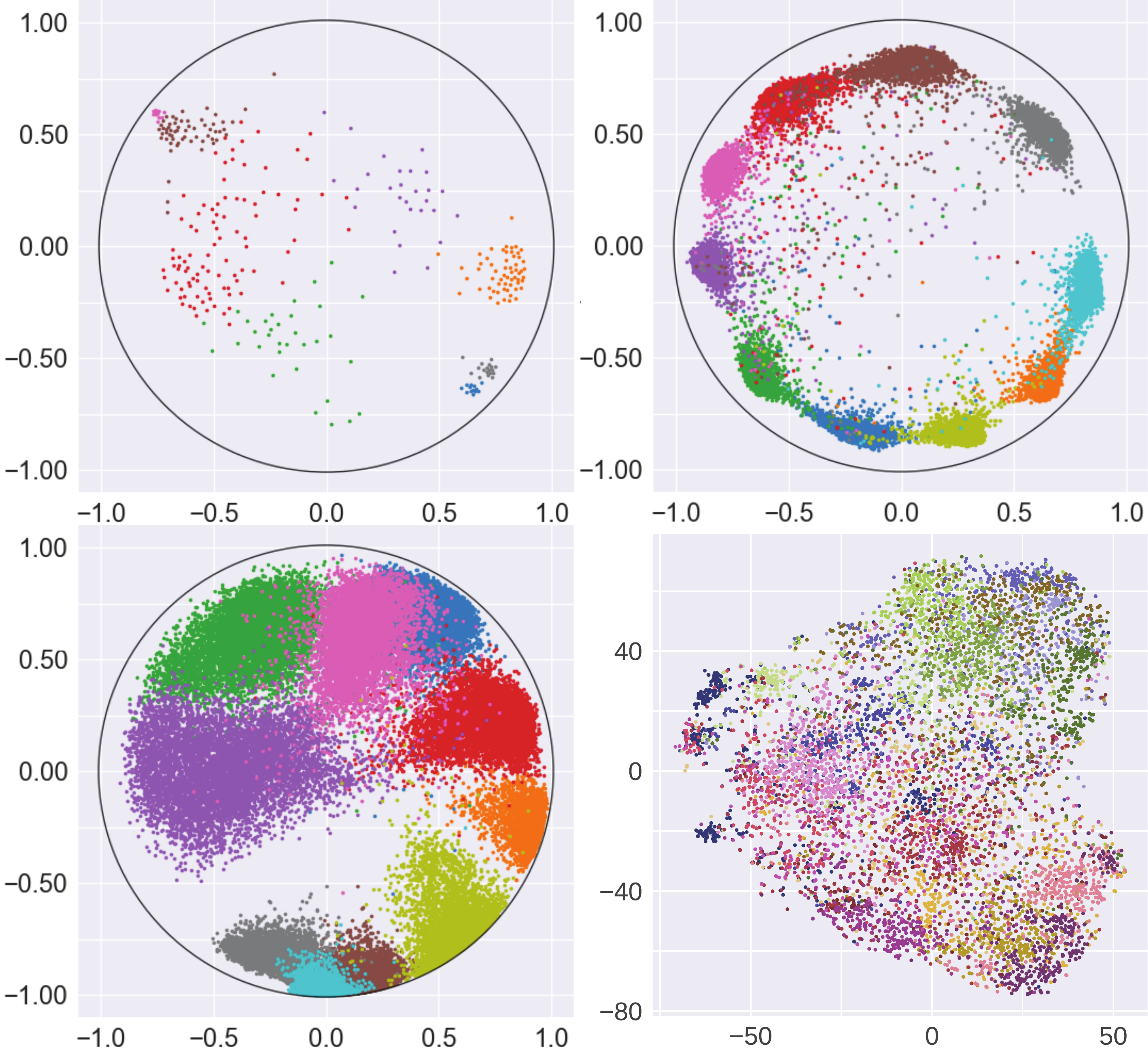}
  \caption{Visualization of four embedded data sets: Olsson's single-cell RNA expression data (top left, $K=8, d=2$), CIFAR10 (top right, $K=10, d=2$), Fashion-MNIST (bottom left, $K=10, d=2$) and mini-ImageNet (bottom right, $K=20, d=512$). Here $K$ stands for the number of classes and $d$ stands for the dimension of embedded Poincar\'e ball. Data points from mini-ImageNet are mapped into $2$ dimensions using tSNE for viewing purposes only and thus may not lie in the unit Poincar\'e disk. Different colors represent different classes.}
  \label{fig:real_data}
\end{figure}

We give an affirmative answer to this question for a wide variety of classification algorithms. By redefining the notion of separation hyperplanes in hyperbolic spaces, we describe the first known Poincar\'e ball perceptron, second-order perceptron and SVM methods with provable performance guarantees. Our perceptron algorithm resolves convergence problems associated with the perceptron method in~\cite{weber2020robust} while the second algorithms generalize the work~\cite{cesa2005second} on second-order perceptrons in Euclidean spaces. Both methods are of importance as they represent a form of online learning in hyperbolic spaces and are basic components of hyperbolic neural networks. On the other hand, our Poincar\'e SVM method successfully addresses issues associated with solving and analyzing a nontrivial nonconvex optimization problem used to formulate hyperboloid SVMs in~\cite{cho2019large}. In the latter case, a global optimum may not be attainable using projected gradient descent methods and consequently this SVM method does not provide tangible guarantees. Our proposed algorithms may be viewed as ``shallow'' one-layer neural networks for hyperbolic spaces that are not only scalable but also (unlike deep networks~\cite{ganea2018hyperbolic,shimizu2020hyperbolic}) exhibit an extremely small storage-footprint. They are also of significant relevance for few-shot meta-learning~\cite{lee2019meta} and for applications such as single-cell subtyping and image data processing as described in our experimental analysis (see Figure~\ref{fig:real_data} for the Poincar\'e embedding of these data sets). 

For our algorithmic solutions we choose to work with the Poincar\'e ball model for several practical and mathematical reasons. First, this model lends itself to ease of data visualization and it is known to be conformal. Furthermore, many recent deep learning models are designed to operate on the Poincar\'e ball model, and our detailed analysis and experimental evaluation of the perceptron, SVM and related algorithms can improve our understanding of these learning methods. The key insight is that \emph{tangent spaces} of points in the Poincar\'e ball model are Euclidean. This, along with the fact that logarithmic and exponential maps are readily available to switch between different relevant spaces simplifies otherwise complicated derivations and allows for addressing classification tasks in a unified manner using convex programs. In our proofs we also use new convex hull algorithms over the Poincar\'e ball model and explain how to select free parameters of the classifiers.

\begin{figure*}[!t]
  \centering
  \subfigure[Accuracy vs $d$]{\includegraphics[trim={0cm 0cm 0cm 0cm},clip,width=0.24\linewidth]{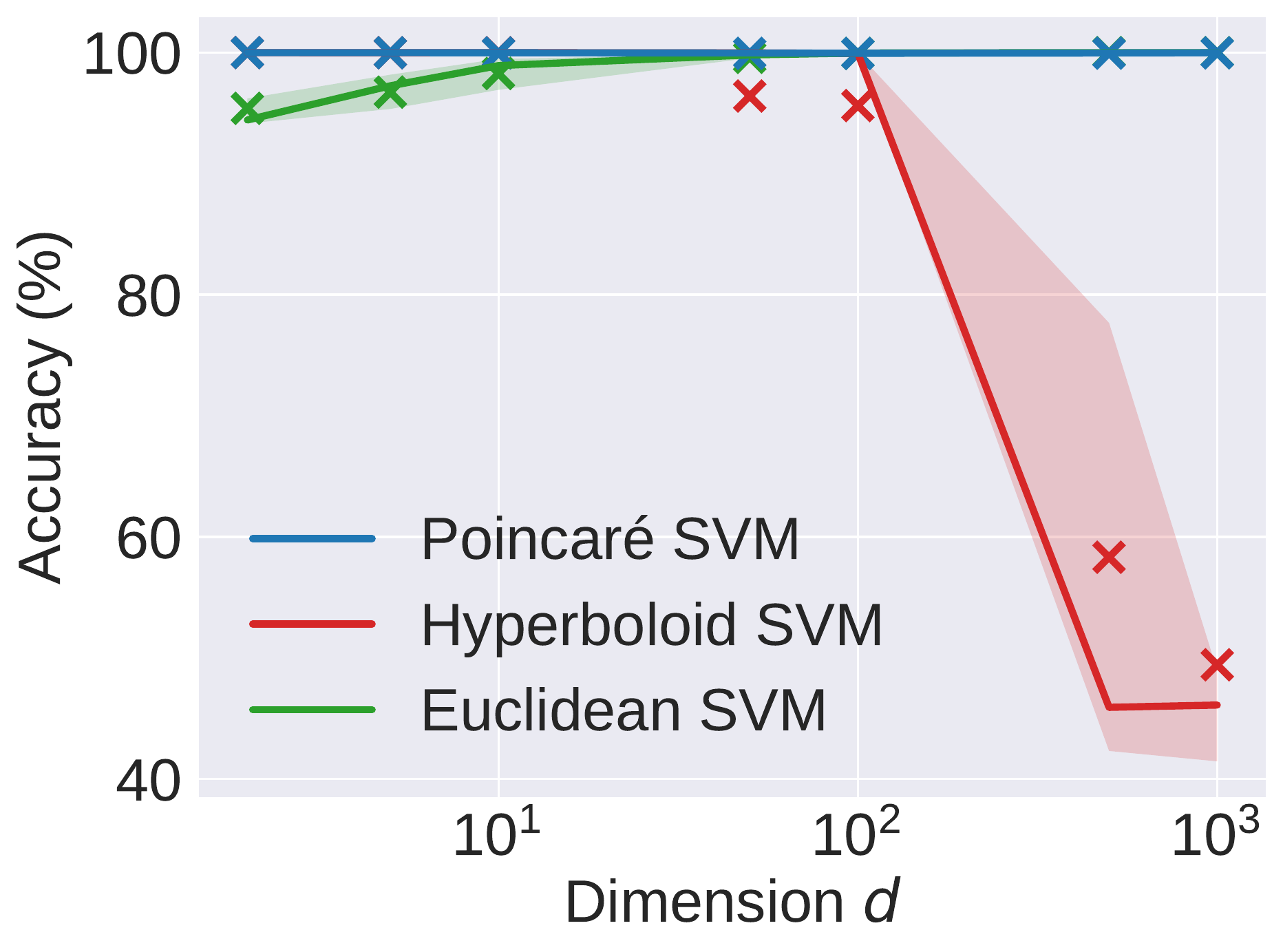}}
  \subfigure[Time vs $d$]{\includegraphics[trim={0cm 0cm 0cm 0cm},clip,width=0.24\linewidth]{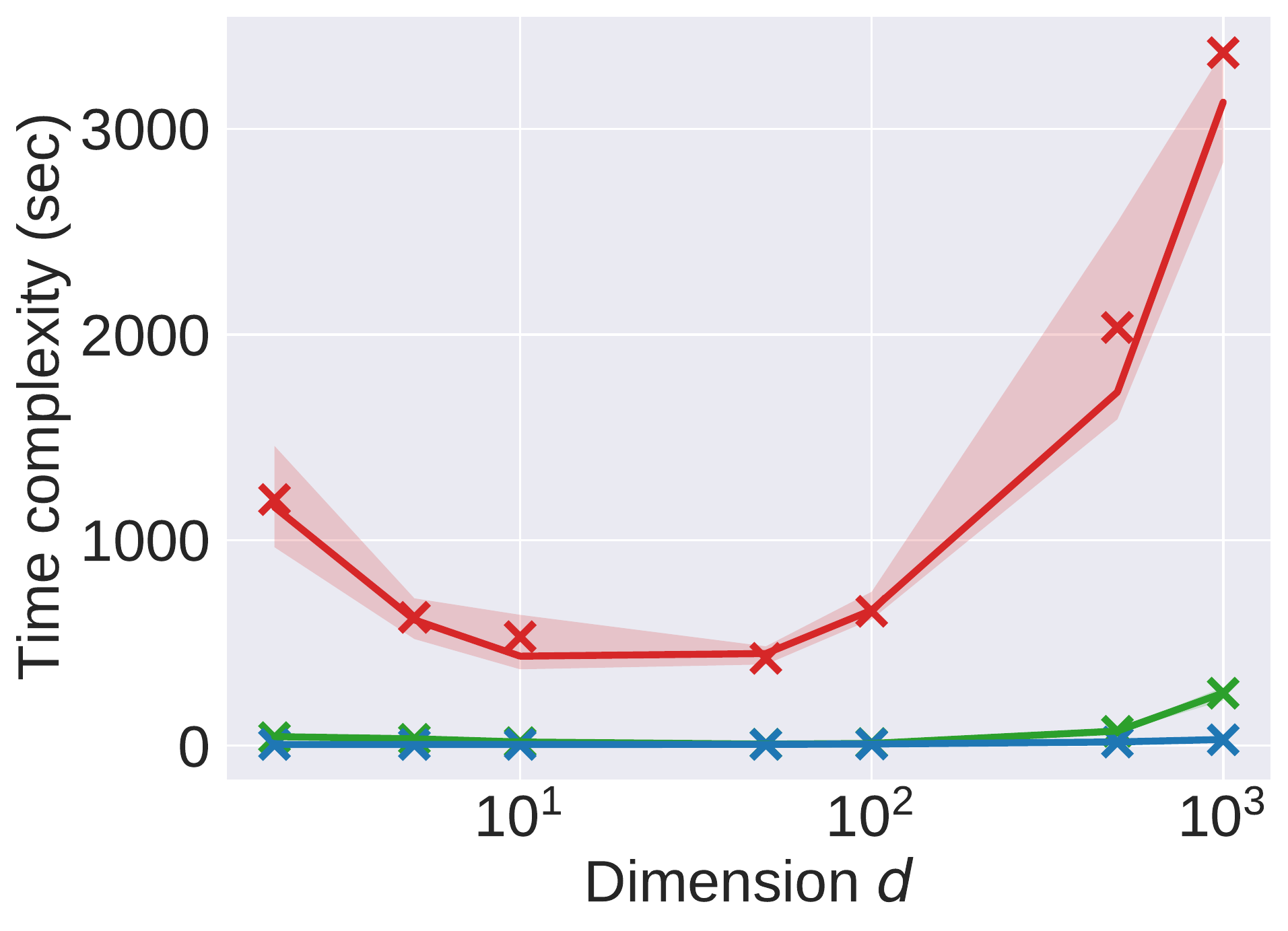}}
  \subfigure[Accuracy vs $N$]{\includegraphics[trim={0cm 0cm 0cm 0cm},clip,width=0.24\linewidth]{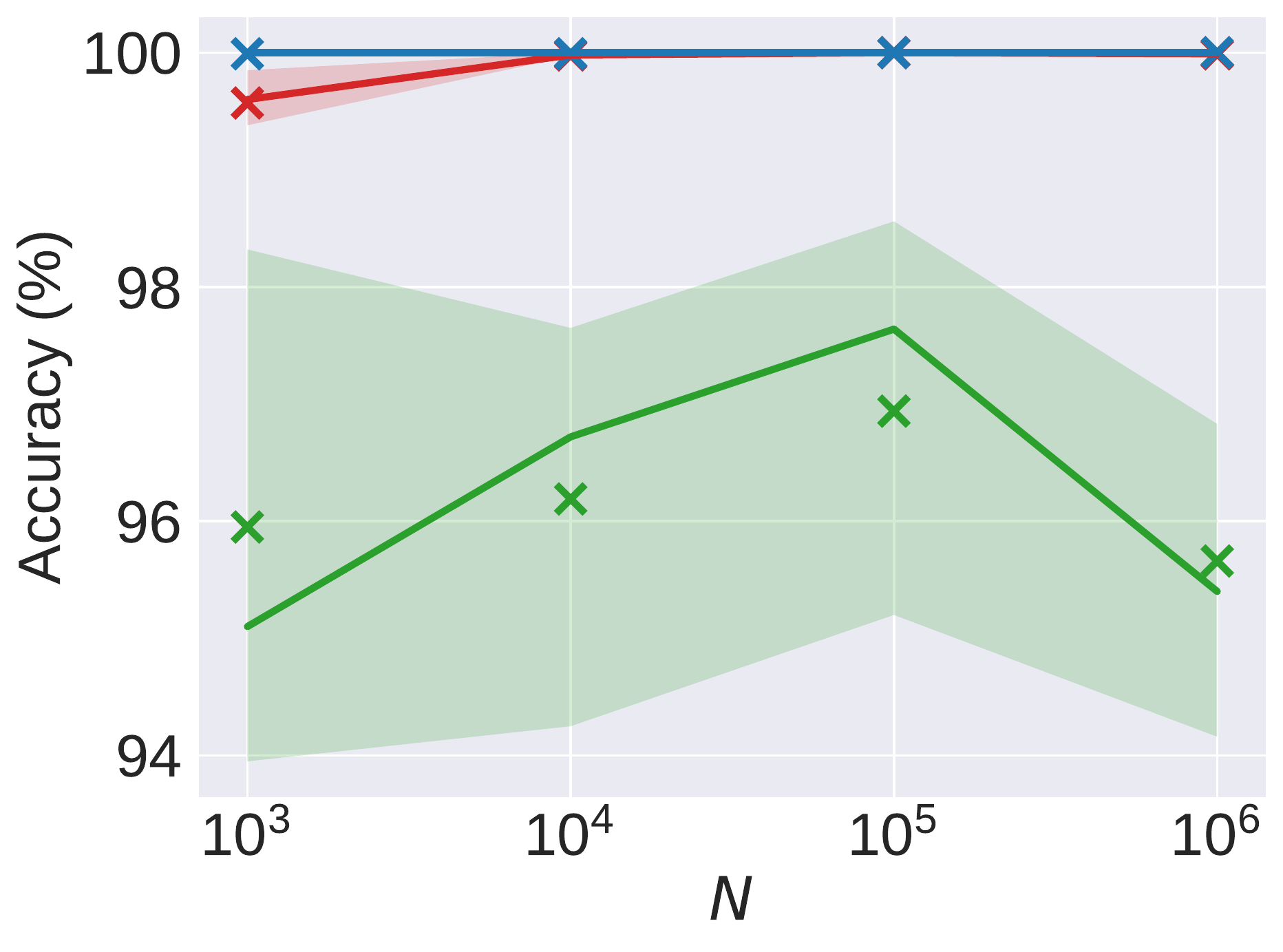}}
  \subfigure[Time vs $N$]{\includegraphics[trim={0cm 0cm 0cm 0cm},clip,width=0.24\linewidth]{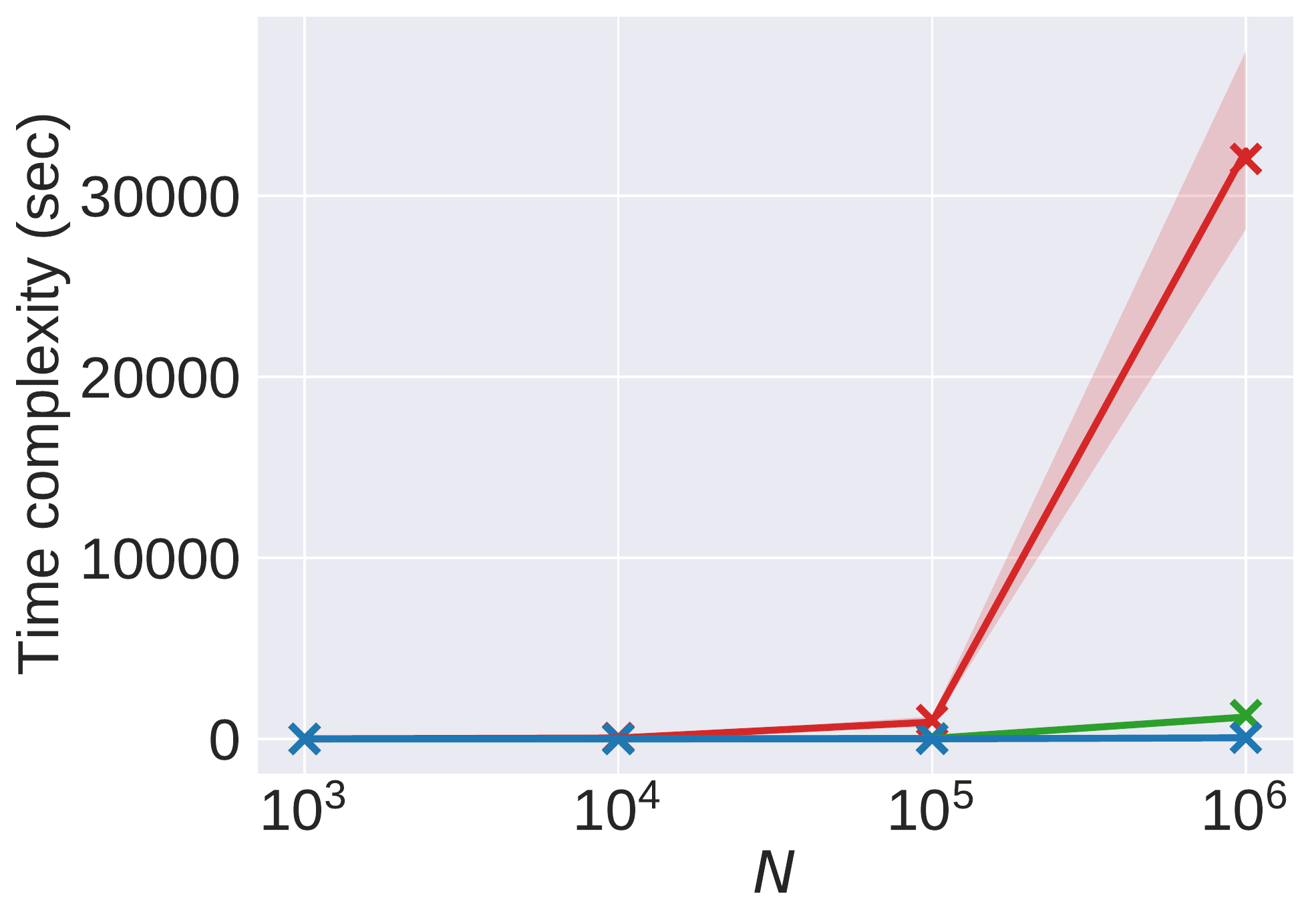}}
  \caption{Classification of $N$ points in $d$ dimensions selected uniformly at random in the Poincar\'e ball. The upper and lower boundaries of the shaded region represent the first and third quantile, respectively. The line shows the medium (second quantile) and the marker $\times$ indicates the mean. Detailed explanations pertaining to the test results can be found in Section~\ref{sec:experiments}}\label{fig:motivation}
\end{figure*}


The proposed Poincar\'e perceptron, second-order perceptron and SVM method easily operate on massive synthetic data sets comprising millions of points and up to one thousand dimensions. Both the Poincar\'e perceptron and its second-order variant converge to an error-free result provided that the data satisfies an $\varepsilon$-margin assumption. The second-order Poincar\'e perceptron converges using significantly fewer iterations than the perceptron method, which matches the advantages offered by its Euclidean counterpart~\cite{cesa2005second}. This is of particular interest in online learning settings and it also results in lower excess risk~\cite{cesa2004generalization}. Our Poincar\'e SVM formulation, which unlike~\cite{cho2019large} comes with provable performance guarantees, also operates significantly faster than its nonconvex counterpart ($1$ minute versus $9$ hours on a set of $10^6$ points, which is $540\times$ faster) and also offers improved classification accuracy as high as $50$\%. Real-world data experiments involve single-cell RNA expression measurements~\cite{olsson2016single}, CIFAR10~\cite{krizhevsky2009learning}, Fashion-MNIST~\cite{xiao2017fashion} and mini-ImageNet~\cite{ravi2016optimization}. These data sets have challenging overlapping-class structures that are hard to process in Euclidean spaces, while Poincar\'e SVMs still offer outstanding classification accuracy with gains as high as $47.91\%$ compared to their Euclidean counterparts.

The paper is organized as follows. Section~\ref{sec:relevance} describes an initial set of experimental results illustrating the scalability and high-quality performance of our SVM method compared to the corresponding Euclidean and other hyperbolic classifiers. This section also contains a discussion 
of prior works on hyperbolic perceptrons and SVMs that do not use the tangent space formalism and hence fail to converge and/or provide provable convergence guarantees. Section~\ref{sec:analysis} contains a review of relevant concepts from differential geometry needed to describe the classifiers as well as our main results, analytical convergence guarantees for the proposed learners. A more detailed set of experimental results, pertaining to real-world single-cell RNA expression measurements for cell-typing and three collections of image data sets is presented in Section~\ref{sec:experiments}. These results illustrate the expressional power of hyperbolic spaces for hierarchical data and highlight the unique feature and performance of our techniques. 

\section{Relevance and Related Work} \label{sec:relevance}
To motivate the need for new classification methods in hyperbolic spaces we start by presenting illustrative numerical results for synthetic data sets. We compare the performance of our Poincar\'e SVM with the previously proposed hyperboloid SVM~\cite{cho2019large} and Euclidean SVM. The hyperbolic perceptron outlined in~\cite{weber2020robust} does not converge and is hence not used in our comparative study. Rigorous descriptions of all mathematical concepts and pertinent proofs are postponed to the next sections and/or the full version of this paper.

One can clearly observe from Figure~\ref{fig:motivation} that the accuracy of Euclidean SVMs may be significantly below $100\%$, as the data points are not linearly separable in Euclidean but rather only in the hyperbolic space. Furthermore, the nonconvex SVM method of~\cite{cho2019large} does not scale well as the number of points increases: It takes roughly $9$ hours to complete the classification process on $10^6$ points while our Poincar\'e SVM takes only $1$ minute. Furthermore, the algorithm breaks down when the data dimension increases to $1,000$ due to its intrinsic non-stability. Only our Poincar\'e SVM can achieve nearly optimal ($100\%$) classification accuracy with extremely low time complexity for all data sets considered. More extensive experimental results on synthetic and real-world data can be found in Section~\ref{sec:experiments}.

The exposition in our subsequent sections explains what makes our classifiers as fast and accurate as demonstrated, especially when compared to the handful of other existing hyperbolic space methods. In the first line of work to address SVMs in hyperbolic spaces~\cite{cho2019large} the authors chose to work with the hyperboloid model of hyperbolic spaces which resulted in a nonconvex optimization problem formulation. The nonconvex problem was solved via projected gradient descent which is known to be able to provably find only a \emph{local} optimum. In contrast, as we will show, our Poincar\'e SVM \emph{provably} converges to a global optimum. The second related line of work~\cite{weber2020robust} studied hyperbolic perceptrons and a hyperbolic version of robust large-margin classifiers for which a performance analysis was included. This work also solely focused on the hyperboloid model and the hyperbolic perceptron method outlined therein does not converge. Since we choose to work with the Poincar\'e ball instead of the hyperboloid model, we can resort to straightforward, universal and simple proof techniques that ``transfer'' the classification problem back from the Poincar\'e to the Euclidean space through the use of tangent spaces. Our analytical convergence results are extensively validated experimentally.

In addition to the two linear classification procedures described above, a number of hyperbolic neural networks solutions have been put forward as well~\cite{ganea2018hyperbolic,shimizu2020hyperbolic}. These networks were built upon the idea of Poincar\'e hyperplanes and motivated our approach for designing Poincar\'e-type perceptrons and SVMs. One should also point out that there are several other deep learning methods specifically designed for the Poincar\'e ball model, including hyperbolic graph neural networks~\cite{liu2019hyperbolic} and Variational Autoencoders~\cite{nagano2019wrapped,mathieu2019continuous,skopek2019mixed}. Despite the excellent empirical performance of these methods theoretical guarantees are still unavailable due to the complex formalism of deep learners. Our algorithms and proof techniques illustrate for the first time why elementary components of such networks, such as perceptrons, perform exceptionally well when properly formulated for a Poincar\'e ball. 

\section{Classification in Hyperbolic Spaces} \label{sec:analysis}

We start with a review of basic notions pertinent to hyperbolic spaces. We then proceed to introduce the notion of \emph{separation hyperplanes} in the Poincar\'e model of hyperbolic space which is crucial for all our subsequent derivations.

\textbf{The Poincar\'e ball model. } Despite the existence of a multitude of equivalent models for hyperbolic spaces, Poincar\'e ball models have received the broadest attention in the machine learning and data mining communities.  This is due to the fact that the Poincar\'e ball model provides conformal representations of shapes and point sets, i.e., in other words, it preserves Euclidean angles of shapes. The model has also been successfully used for designing hyperbolic neural networks~\cite{ganea2018hyperbolic,shimizu2020hyperbolic} with excellent heuristic performance. Nevertheless, the field of learning in hyperbolic spaces -- under the Poincar\'e or other models  -- still remains largely unexplored.

The Poincar\'e ball model $(\mathbb{B}^n_c,g^{\mathbb{B}})$ is a Riemannian manifold. For the absolute value of the curvature $c>0$, its domain is the open ball of radius $1/\sqrt{c}$, i.e., $\mathbb{B}^n_c=\{x\in \mathbb{R}^n:\sqrt{c}\,\|x\|< 1\}$. Here and elsewhere $\|\cdot\|$ stands for the $\ell_2$ norm and $\ip{\cdot}{\cdot}$ stands for the standard inner product.  The Riemannian metric is defined as $g^{\mathbb{B}_c}_x(\cdot,\cdot) = (\sigma_x^c)^2\ip{\cdot}{\cdot}$, where $\sigma_x^c = 2/(1-c\|x\|^2)$. For $c=0$, we recover the Euclidean space, i.e., $\mathbb{B}^n_c = \mathbb{R}^n$. For simplicity, we focus on the case $c=1$ albeit our results can be generalized to hold for arbitrary $c>0$. Furthermore, for a reference point $p\in \mathbb{B}^n$, we denote its tangent space, the first order linear approximation of $\mathbb{B}^n$ around $p$, by $T_p\mathbb{B}^n$.

In the following, we introduce M\"obius addition and scalar multiplication --- two basic operators on the Poincar\'e ball~\cite{ungar2008analytic}. These operators represent analogues of vector addition and scalar-vector multiplication in Euclidean spaces. The M\"obius addition of $x,y\in\mathbb{B}^n$ is defined as
\begin{align}\label{eq:Mobius_add}
    & x \oplus y = \frac{(1+2\ip{x}{y}+\|y\|^2)x + (1-\|x\|^2)y}{1+2\ip{x}{y}+\|x\|^2\|y\|^2}.
\end{align}
Unlike its vector-space counterpart, this addition is noncommutative and nonassociative. The M\"obius version of multiplication of $x\in \mathbb{B}^n\setminus\{0\}$ by a scalar $r\in \mathbb{R}$ is defined according to
\begin{align}\label{eq:Mobius_mul}
    & r \otimes x = \tanh(r\tanh^{-1}(\|x\|))\frac{x}{\|x\|} \text{ and }r\otimes 0 = 0.
\end{align}
For detailed properties of these operations, see~\cite{vermeer2005geometric,ganea2018hyperbolic}. The distance function in the Poincar\'e model is 
\begin{align}
     d(x,y) = 2\tanh^{-1}(\|(-x)\oplus y\|).
\end{align}
Using M\"obius operations one can also describe geodesics (analogues of straight lines in Euclidean spaces) in $\mathbb{B}^n$. The geodesics connecting two points $x,y\in \mathbb{B}^n$ is given by
\begin{align}\label{eq:geodesic}
    \gamma_{x\rightarrow y}(t) = x \oplus(t\otimes((-x)\oplus y)).
\end{align}
Note that $t \in [0,1]$ and $\gamma_{x\rightarrow y}(0) = x$ and $\gamma_{x\rightarrow y}(1) = y$. 


The following result explains how to construct a geodesic with a given starting point and a tangent vector.
\begin{lemma}[\cite{ganea2018hyperbolic}]
 For any $p\in \mathbb{B}^n$ and $v\in T_p\mathbb{B}^n$ s.t. $g_p(v,v)=1$, the geodesic starting at $p$ with tangent vector $v$ equals:
 \begin{align*}
     & \gamma_{p,v}(t) = p \oplus \left(\tanh\left(\frac{t}{2}\right)\frac{v}{\|v\|} \right),\\
     &\text{where }\gamma_{p,v}(0)=p \text{ and }\dot{\gamma}_{p,v}(0) = v.
 \end{align*}
\end{lemma}
We complete the overview by introducing logarithmic and exponential maps.
\begin{lemma}[Lemma 2 in~\cite{ganea2018hyperbolic}]
 For any point $p\in\mathbb{B}^n$ the exponential map $\exp_p:T_p\mathbb{B}^n\mapsto \mathbb{B}^n$ and the logarithmic map $\log_p: \mathbb{B}^n\mapsto T_p\mathbb{B}^n$ are given for $v\neq 0$ and $x\neq p$ by:
 \begin{align}\label{eq:explogmap}
     & \exp_p(v) = p \oplus \left(\tanh\left(\frac{\sigma_p\|v\|}{2}\right)\frac{v}{\|v\|}\right),\\
     & \log_p(x) = \frac{2}{\sigma_p}\tanh^{-1}(\|(-p)\oplus x\|)\frac{(-p)\oplus x}{\|(-p)\oplus x\|}.
 \end{align}
\end{lemma}
The geometric interpretation of $\log_p(x)$ is that it gives the tangent vector $v$ of $x$ for starting point $p$. On the other hand, $\exp_p(v)$ returns the destination point $x$ if one starts at the point $p$ with tangent vector $v$. Hence, a geodesic from $p$ to $x$ may be written as
\begin{align}
    \gamma_{p\rightarrow x}(t) = \exp_p(t\log_p(x)),\quad t\in [0,1].
\end{align}
See Figure~\ref{fig:1} for the illustration.

\textbf{Classification with Poincar\'e hyperplanes.} The recent work~\cite{ganea2018hyperbolic} introduced the notion of a Poincar\'e hyperplane which generalizes the concept of a hyperplane in Euclidean space. The Poincar\'e hyperplane with reference point $p\in\mathbb{B}^n$ and normal vector $w\in T_p\mathbb{B}^n$ in the above context is defined as
\begin{align*}
    &H_{w,p} = \{x\in \mathbb{B}^n: \ip{\log_p(x)}{w}_p=0\}\\
    &=\{x\in \mathbb{B}^n: \ip{(-p)\oplus x}{w}=0\},
\end{align*}
where $\ip{x}{y}_p = (\sigma_p)^2\ip{x}{y}$. The minimum distance of a point $x\in \mathbb{B}^n$ to $H_{w,p}$ has the following close form
\begin{align}\label{eq:p2H_dist}
    & d(x,H_{w,p}) = \sinh^{-1}\left(\frac{2|\langle (-p)\oplus x, w\rangle|}{(1-\|(-p)\oplus x\|^2)\|w\|}\right).
\end{align}
We find it useful to restate~\eqref{eq:p2H_dist} so that it only depends on vectors in the tangent space $T_p\mathbb{B}^n$ as follows.
\begin{lemma}\label{lma:p2H_dist2}
    Let $v = \log_p(x)$ (and thus $x = \exp_p(v)$), then we have
    \begin{align}\label{eq:p2H_dist2}
        & d(x,H_{w,p}) = \sinh^{-1}\left(\frac{2\tanh(\frac{\sigma_p\|v\|}{2})|\langle v, w\rangle|}{(1-\tanh(\frac{\sigma_p\|v\|}{2})^2)\|w\|\|v\|}\right)
    \end{align}
\end{lemma}

Equipped with the above definitions, we now focus on binary classification in Poincar\'e models. To this end, 
let $\{(x_i,y_i)\}_{i=1}^N$ be a set of $N$ data points, where $x_i\in \mathbb{B}^n$ and $y_i\in\{\pm 1\}$ represent the true labels. Note that based on Lemma~\ref{lma:p2H_dist2}, the decision function based on $H_{w,p}$ is $h_{w,p}(x) = \text{sign}\left(\ip{\log_p(x_i)}{w}\right)$. This is due to the fact that $\sinh^{-1}$ does not change the sign of its input and that all other terms in~\eqref{eq:p2H_dist2} are positive if $v_i=\log_p(x_i)$ and $w \neq 0$. For the case that either $\log_p(x_i)$ or $w$ is $0$, $\ip{v}{w} = 0$ and thus the sign remains unchanged. For linear classification, the goal is to learn $w$ that correctly classifies all points. For large margin classification, we further required that the learnt $w$ achieves the largest possible margin, 
\begin{align}
    \max_{w\in T_p\mathbb{B}^n} \min_{i\in [N]} y_i h_{w,p}(x_i) d(x_i,H_{w,p}).
\end{align}

\subsection{Classification algorithms for Poincar\'e balls}\label{sec:GA2P}

\begin{figure}[!t]
  \centering
  \subfigure[Poincar\'e disk]{\includegraphics[trim={9cm 3cm 9cm 3cm},clip,width=0.49\linewidth]{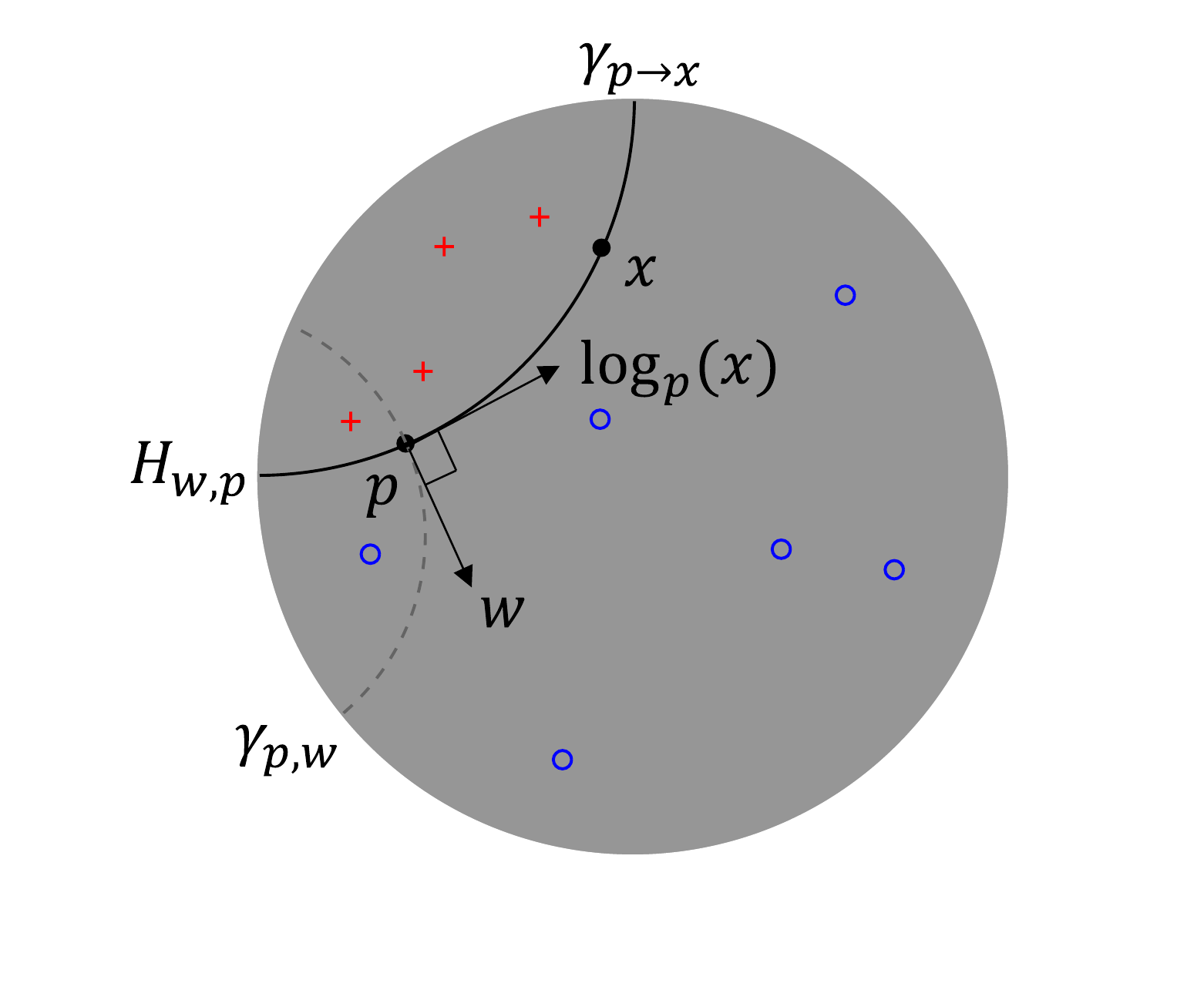}}
  \subfigure[Tangent space]{\includegraphics[trim={9cm 6cm 9cm 3cm},clip,width=0.49\linewidth]{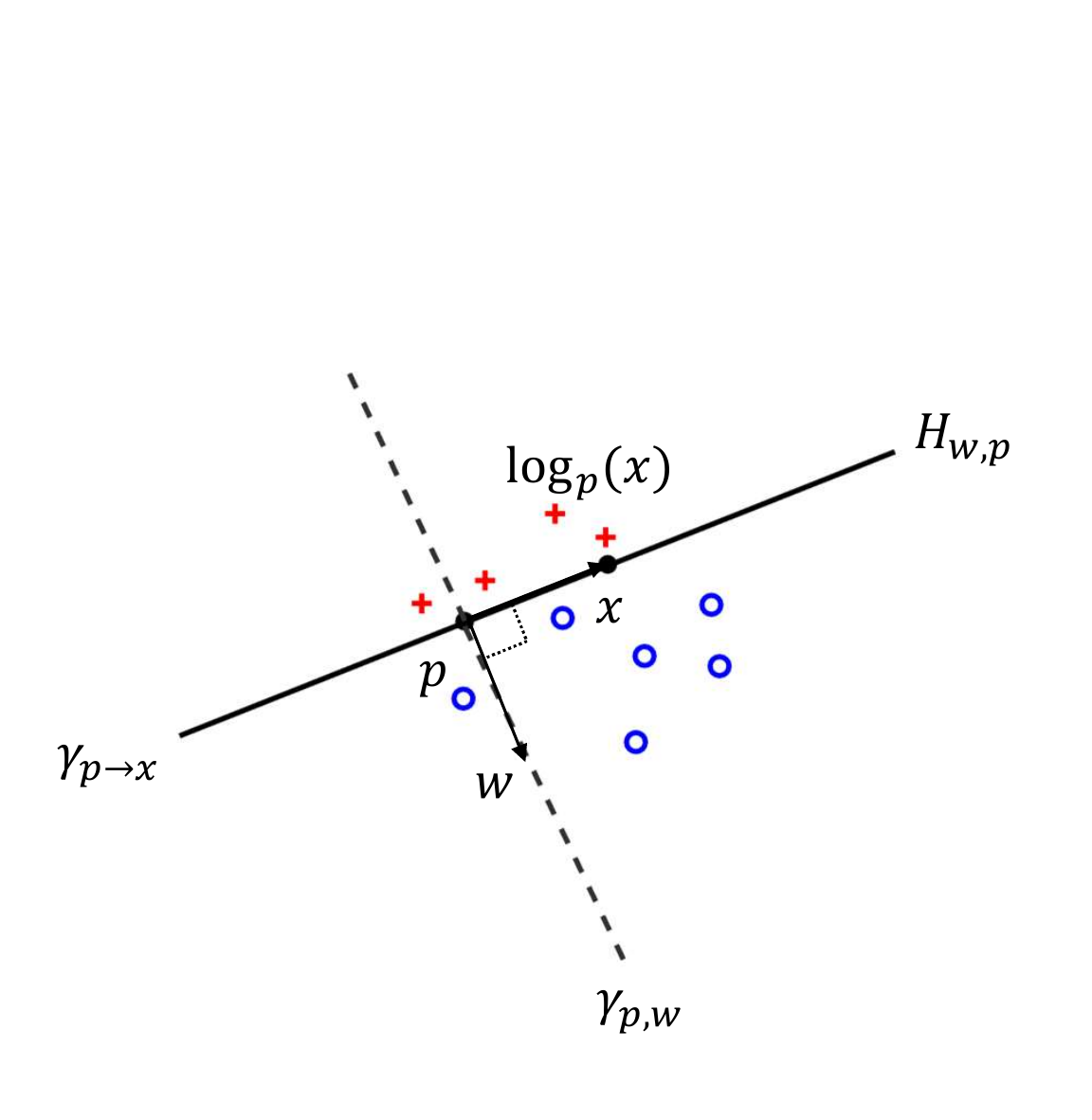}}
  \caption{Figure (a): A linear classifier in Poincar\'e disk $\mathbb{B}^2$. Figure (b): Corresponding tangent space $T_p\mathbb{B}^2$.}\label{fig:1}
\end{figure}

In what follows we outline the key idea behind our approach to classification and analysis of the underlying algorithms. We start with the perceptron classifier, which is the simplest approach yet of relevance for online settings. We then proceed to describe its second-order extension which offers significant reductions in the number of data passes and then introduce our SVM method which offers excellent performance with provable guarantees.

Our approach builds upon the result of Lemma~\ref{lma:p2H_dist2}. For each $x_i\in \mathbb{B}^n$, let $v_i = \log_p(x_i)$. We assign a corresponding weight as
\begin{align}\label{eq:eta}
    \eta_i = \frac{2\tanh(\frac{\sigma_p\|v_i\|}{2})}{(1-\tanh(\frac{\sigma_p\|v_i\|}{2})^2)\|v_i\|}.
\end{align}
Without loss of generality, we also assume that the optimal normal vector has unit norm $\|w^\star\|=1$. 
Then~\eqref{eq:p2H_dist2} can be rewritten as
\begin{align}\label{eq:simp_p2H_tp}
    d(x_i,H_{w^\star,p}) = \sinh^{-1}\left( \eta_i|\ip{v_i}{w^\star}| \right).
\end{align}
Note that $\eta_i \geq 0$ and $\eta_i=0$ if and only if $v_i=0$, which corresponds to the case $x_i=p$. Nevertheless, this ``border'' case can be easily eliminated under a margin assumption. Hence, the problem of finding an optimal classifier becomes similar to the Euclidean case if one focuses on the \emph{tangent space} of the Poincar\'e ball model (see Figure~\ref{fig:1} for an illustration).

\subsection{Poincar\'e perceptron}
We first restate the standard assumptions needed for the analysis of the perceptron algorithm in Euclidean space for the Poincar\'e model.
\begin{assumption}\label{asp:all}
    \begin{align}
        & \exists w^\star \in T_p \mathbb{B}^n,\;y_i\ip{\log_p(x_i)}{w^\star}>0\;\forall i\in[N],\label{asp:1}\\
        & \exists \epsilon>0\;\text{s.t.}\;d(x_i,H_{w^\star,p})\geq \varepsilon\;\forall i\in[N],\label{asp:2}\\
        & \|x_i\| \leq R < 1\;\forall i\in [N]. \label{asp:3}
    \end{align}
\end{assumption}
The first assumption~\eqref{asp:1} postulates the existence of a classifier that correctly classifies every points. The margin assumption is listed in~\eqref{asp:2}, while \eqref{asp:3} ensures that points lie in a bounded region.

Using~\eqref{eq:simp_p2H_tp} we can easily design the Poincar\'e perceptron update rule. If the $k^{th}$ mistake happens at instance $(x_{i_k},y_{i_k})$ (i.e., $y_{i_k}\ip{\log_p(x_{i_k})}{w_k}\leq 0$), then
\begin{align}\label{eq:PP}
    & w_{k+1} = w_k + \eta_{i_k}y_{i_k}\log_p(x_{i_k}),\;w_1=0.
\end{align}

The Poincar\'e perceptron algorithm~\eqref{eq:PP} comes with the following convergence guarantees.
\begin{theorem}\label{thm:HPbound_p}
Under Assumption~\ref{asp:all}, the Poincar\'e perceptron~\eqref{eq:PP} will correctly classify all points with at most $\left(\frac{2R_p}{(1-R_p^2)\sinh(\varepsilon)}\right)^2$ updates, where $R_p = \frac{\|p\|+R}{1+\|p\|R}$.
\end{theorem}

To prove Theorem~\ref{thm:HPbound_p}, we need the technical lemma below. 
\begin{lemma}\label{lma:paramax_mobadd}
Let $a\in\mathbb{B}^n$. Then
    \begin{align}
        &\operatornamewithlimits{argmax}_{b\in \mathbb{B}^n:\|b\|\leq R} \|a\oplus b \| = R\frac{a}{\|a\|},\\
        &\operatornamewithlimits{argmax}_{b\in \mathbb{B}^n:\|b\|\leq R} \|b\oplus a \| = R\frac{a}{\|a\|}.
    \end{align}
\end{lemma}
If we replace $\oplus$ with ordinary addition then we can basically interpret the result as follows: The norm of $\|a+b\|$ is maximized when $a$ has the same direction as $b$. This can be easily proved by invoking the Cauchy-Schwartz inequality. However, it is nontrivial to show the result under M\"obius addition.

\textbf{proof. }As already mentioned in Section~\ref{sec:GA2P}, the key idea is to work in the tangent space $T_p\mathbb{B}^n$, in which case the Poincar\'e perceptron becomes similar to the Euclidean perceptron. First, we establish the boundedness of the tangent vectors $v_i=\log_p(x_i)$ in $T_p\mathbb{B}^n$ by invoking the definition of $\sigma_p$ from Section~\ref{sec:analysis}:
\begin{align}\label{eq:thm1_pfeq1}
    & \|v_i\|=\|\log_p(x_i)\| = \frac{2}{\sigma_p}\tanh^{-1}(\|(-p)\oplus x_i\|)\nonumber\\
    & \stackrel{(a)}{\leq}\frac{2}{\sigma_p}\tanh^{-1}\left(R_p\right)\Leftrightarrow \tanh(\frac{\sigma_p\|v_i\|}{2})\leq R_p,
\end{align}
where (a) can be shown to hold true by involving Lemma~\ref{lma:paramax_mobadd} and performing some algebraic manipulations. Details are as follows.
\begin{align*}
    & \|(-p)\oplus x_i\| \leq \|(-p)\oplus \frac{R}{\|p\|}(-p))\| \tag*{By Lemma~\ref{lma:paramax_mobadd}}\\
    & = \left\|\frac{(1+2R\|p\|+R^2)(-p)+(1-\|p\|^2)\frac{R(-p)}{\|p\|}}{(R+\|p\|^2)^2}\right\| \tag*{By~\eqref{eq:Mobius_add}}\\
    & = \frac{\|p\|+R^2\|p\|+R\|p\|^2+R}{(R+\|p\|^2)^2} = \frac{(R+\|p\|)(1+R\|p\|)}{(R+\|p\|^2)^2}\\
    & = \frac{1+R\|p\|}{R+\|p\|^2} = R_p.
\end{align*}
Combining with the fact that $tanh^{-1}(\cdot)$ is non-decreasing in $[0,1)$, we arrive the inequality (a).

The remainder of the analysis is similar to that of the standard Euclidean perceptron. We first lower bound $\|w_{k+1}\|$ as 
\begin{align}\label{eq:thm1_pfeq2}
    & \|w_{k+1}\| \stackrel{(b)}{\geq} \ip{w_{k+1}}{w^\star} = \ip{w_k}{w^\star} + \eta_{i_k}y_{i_k}\ip{v_{i_k}}{w^\star}\nonumber\\
    &\stackrel{(c)}{\geq} \ip{w_k}{w^\star} + \sinh(\varepsilon) \geq \cdots \geq k\sinh(\varepsilon),
\end{align}
where (b) follows from the Cauchy-Schwartz inequality and (c) is a consequence of the margin assumption~\eqref{asp:2}. Next, we upper bound $\|w_{k+1}\|$ as 
\begin{align}\label{eq:thm1_pfeq3}
    &\|w_{k+1}\|^2 \leq \|w_k\|^2 + \left(\frac{2\tanh(\frac{\sigma_p\|v_{i_k}\|}{2})}{(1-\tanh(\frac{\sigma_p\|v_{i_k}\|}{2})^2)}\right)^2 \nonumber \\
    & \stackrel{(d)}{\leq} \|w_k\|^2 + \left(\frac{2R_p}{(1-R_p^2)}\right)^2 \leq \cdots \leq k\left(\frac{2R_p}{(1-R_p^2)}\right)^2,
\end{align}
where (d) is a consequence of~\eqref{eq:thm1_pfeq1} and the fact that the function $f(x) = \frac{x}{1-x^2}$ is nondecreasing for $x\in(0,1)$. Note that $R_p < 1$ since $\|p\|<1,R<1$ and $0<(1-\|p\|)(1-R) = 1+\|p\|R -(\|p\|+R).$
Combining~\eqref{eq:thm1_pfeq2} and~\eqref{eq:thm1_pfeq3} we complete the proof.

It is worth pointing out that the authors of~\cite{weber2020robust} also designed and analyzed a different version of a hyperbolic perceptron in the hyperboloid model $\mathbb{L}^d = \{x\in \mathbb{R}^{d+1}:[x,x] = -1\}$, where $[x,y] = x^THy,H=diag(-1,1,1,\ldots,1)$ denotes the Minkowski product. Their proposed update rule is
\begin{align}
    & u_{k} = w_{k} + y_{n} x_{n} \ \mbox{ if }  -y_n [w_{k}, x_n] < 0 \label{eq:WebP_eq1}\\
    & w_{k+1} = u_k/ \min\{1,\sqrt{[u_{k}, u_k]}\}\label{eq:WebP_eq2},
\end{align}
where~\eqref{eq:WebP_eq2} is a normalization step. Although a convergence result was claimed in~\cite{weber2020robust}, we demonstrate by simple counterexamples that their hyperbolic perceptron do not converge. Moreover, we find that the algorithm does not converge to a meaningful solution for most of the data sets tested; this can be easily seen by using the proposed update rule with $w_0=e_2$ and $x_1 = e_1 \in \mathbb{L}^2$ with label $y_1 = 1$, where $e_j$ are standard basis vectors (the particular choice leads to an ill-defined normalization). Other counterexamples $w_0=0$ involve normalization with complex numbers for arbitrary $x_1\in\mathbb{L}^2$ which is not acceptable. 


\subsection{Poincar\'e second-order perceptron}
The reason behind our interest in the second-order perceptron is that it leads to fewer mistakes during training compared to the classical perceptron. It has been shown in~\cite{cesa2004generalization} that the error bounds have corresponding statistical risk bounds in online learning settings, which strongly motivates the use of second-order perceptrons for online classification. The performance improvement of the modified perceptron comes from accounting for second-order data information, as the standard perceptron is essentially a gradient descent (first-order) method.

Equipped with the key idea of our unified analysis, we compute the scaled tangent vectors $z_i = \eta_i v_i = \eta_i \log_p(x_i)$.
Following the same idea, we can extend the second order perceptron to Poincar\'e ball model. Our Poincare second-order perceptron is described in Algorithm~\ref{alg:SOHP} which has the following theoretical guarantee.

\begin{algorithm2e}
\caption{Poincar\'e second order perceptron}\label{alg:SOHP}
\SetAlgoLined
\DontPrintSemicolon
\SetKwInput{Input}{Input}
\SetKwInOut{Output}{Output}
  \Input{Data points $x_i$, labels $y_i$, reference point $p$, parameter $a>0$.
  }
  Initialization: $\xi_0\leftarrow 0$, $X_0 = \emptyset$, $k=1$.\;
  \For {$t=1,2,\ldots$}{
        Get $(x_t,y_t)$, compute $z_t$, set $S_{t} \leftarrow [X_{k-1}\;z_t]$.\;
        Predict $\hat{y}_t = sign(\ip{w_t}{z_t})$, where $w_t=(aI+S_tS_t^T)^{-1}\xi_{k-1}$.\;
        \If{$\hat{y}_i\neq y_i$}{
                    $\xi_k = \xi_{k-1} + y_tz_t$, $X_k\leftarrow S_t$, $k\leftarrow k+1$.\;
                }
        }

\end{algorithm2e}

\begin{theorem}\label{thm:PSOP}
 For all sequences $\left((x_1,y_1),(x_2,y_2),\ldots\right)$ with assumption~\ref{asp:all}, the total number of mistakes $k$ for Poincar\'e second order perceptron satisfies
 \begin{align}
     & k \leq \frac{1}{\sinh(\varepsilon)}\sqrt{(a+\lambda_{w^\star})\left(\sum_{j\in[n]}\log(1+\frac{\lambda_j}{a})\right)},
 \end{align}
 where $\lambda_w = w^T X_kX_k^T w$ and $\lambda_j$ are the eigenvalues of $X_kX_k^T$.
\end{theorem}

The bound in Theorem~\ref{thm:PSOP} has a form that is almost identical to that of its Euclidean counterpart~\cite{cesa2005second}. However, it is important to observe that the geometry of the Poincar\'e ball model plays a important role when evaluating the eigenvalues $\lambda_j$ and $\lambda_w$. Another important observation is that our tangent-space analysis is not restricted to first-order analysis of perceptrons.

\subsection{Poincar\'e SVM}
We conclude our theoretical analysis by describing how to formulate and solve SVMs in the Poincar\'e ball model with performance guarantees. For simplicity, we only consider binary classification. Techniques for dealing with multiple classes are given in Section~\ref{sec:experiments}. 

When data points from two different classes are linearly separable the goal of SVM is to find a ``max-margin hyperplane'' that correctly classifies all data points and has the maximum distance from the nearest point. This is equivalent to selecting two parallel hyperplanes with maximum distance that can separate two classes. Following this approach and assuming that the data points are normalized, we can choose these two hyperplanes as $\langle \log_p(x_i), w\rangle=1$ and $\langle \log_p(x_i), w\rangle=-1$ with $w\in T_p \mathbb{B}^n$. Points lying on these two hyperplanes are referred to as support vectors following the convention for Euclidean SVM. They are critical for the process of selecting $w$.

Let $v_i=\log_p(x_i)\in T_p \mathbb{B}^n$ be such that $|\langle v_i, w\rangle|=1$. Therefore, by Cauchy-Schwartz inequality the support vectors satisfy
\begin{equation}
    \|v_i\|\|w\| \geq |\langle v_i, w\rangle| = 1 \Rightarrow \|v_i\| \geq 1 / \|w\|.
\label{eq:svm_constraint}
\end{equation}
Combing the above result with~(\ref{eq:simp_p2H_tp}) leads to a lower bound on the distance of a data point $x_i$ to the hyperplane $H_{w,p}$
\begin{equation}
    d(x_i,H_{w, p}) \geq \sinh ^{-1}\left(\frac{2 \tanh (\sigma_p/2\|w\|)}{1-\tanh^2(\sigma_p/2\|w\|)}\right),
\label{eq:lower_bound_svm_distance}
\end{equation}
where equality is achieved for $v_i=kw\;(k\in\mathbb{R})$. Thus we can obtain a max-margin classifier in the Poincar\'e ball  that can correctly classify all data points by maximizing the lower bound in~(\ref{eq:lower_bound_svm_distance}). 
Through a sequence of simple reformulations, the optimization problem of maximizing the lower bound~(\ref{eq:lower_bound_svm_distance}) can be cast as an easily-solvable \textbf{convex} problem described in Theorem~\ref{thm:hard-margin-svm-convex}.
\begin{theorem}\label{thm:hard-margin-svm-convex}
Maximizing the margin~(\ref{eq:lower_bound_svm_distance}) is equivalent to solving the convex problem of either primal (P) or dual (D) form:
\begin{align}\label{eq:hard-margin-svm-convex}
    (P)\;&\min_w \frac{1}{2}\|w\|^2\quad \text{s.t.}\ y_i\langle v_i, w\rangle \geq 1\;\forall i\in [N];\\
    (D)\;&\max_{\alpha\geq 0} \sum_{i=1}^N \alpha_i - \frac{1}{2}\left\|\sum_{i=1}^N\alpha_i y_iv_i\right\|^2\;\text{s.t.}\ \sum_{i=1}^N\alpha_i y_i=0,
\end{align}
which is guaranteed to achieve a global optimum with linear convergence rate by stochastic gradient descent.
\end{theorem}


The Poincar\'e SVM formulation from Theorem~\ref{thm:hard-margin-svm-convex} is \emph{inherently different} from the only other known hyperbolic SVM approach~\cite{cho2019large}. There, the problem is nonconvex and thus does not offer convergence guarantees to a global optimum when using projected gradient descent. In contrast, since both (P) and (D) are smooth and strongly convex, variants of stochastic gradient descent is guaranteed to reach a global optimum with a linear convergence rate. This makes the Poincar\'e SVM numerically more stable and scalable to millions of data points. Another advantage of our formulation is that a solution to (D) directly produces the support vectors, i.e., the data points with corresponding $\alpha_i\neq 0$ that are critical for the classification problem.

When two data classes are not linearly separable (i.e., the problem is soft- rather than hard-margin), the goal of the SVM method is to maximize the margin while controlling the number of misclassified data points. Below we define a soft-margin Poincar\'e SVM that trades-off the margin and classification accuracy.

\begin{theorem}\label{thm:soft-margin-svm-convex}
Solving soft-margin Poincar\'e SVM is equivalent to solving the convex problem of either primal (P) or dual (D) form:
\begin{align}\label{eq:soft-margin-svm-convex}
    (P)\;&\min_w \frac{1}{2}\|w\|^2\ + C\sum_{i=1}^N\max\left(0, 1-y_i\langle v_i,w\rangle\right);\\
    (D)\;&\max_{0\leq\alpha\leq C} \sum_{i=1}^N \alpha_i - \frac{1}{2}\left\|\sum_{i=1}^N\alpha_i y_iv_i\right\|^2\;\text{s.t.}\ \sum_{i=1}^n\alpha_i y_i=0,
\end{align}
which is guaranteed to achieve a global optimum with sublinear convergence rate by stochastic gradient descent. 
\end{theorem}

The algorithmic procedure behind the soft-margin Poincar\'e SVM is depicted in Algorithm~\ref{alg:soft-margin-svm}.  

\begin{algorithm2e}
\caption{Soft-margin Poincar\'e SVM}\label{alg:soft-margin-svm}
\SetAlgoLined
\DontPrintSemicolon
\SetKwInput{Input}{Input}
\SetKwInOut{Output}{Output}
  \Input{Data points $x_i$, labels $y_i$, reference point $p$, tolerance $\varepsilon$, maximum iteration number $T$.
  }
  Initialization: $w_0\leftarrow 0, f(w_0) = NC$.\;
  \For {$t=1,2,\ldots$}{
        $f(w_{t})=\frac{1}{2}\|w_t\|^2\ + C\sum_{i=1}^N\max\left(0, 1-y_i\langle v_i,w_t\rangle\right)$.\;
        \If{$f(w_{t})$ - $f(w_{t-1})\leq \varepsilon$ or $t>T$}{
                    \textbf{break};
                }
        Sample $(x_t, y_t)$ from data set randomly and compute $v_t=\log_p(x_t)$.\;
        \uIf{$1-y_t\langle v_t,w_{t-1}\rangle < 0$}{
                    $\nabla w = w_{t-1}$.\;
                }
        \Else{$\nabla w = w_{t-1} - NCy_tv_t$.\;}
        $w_t = w_{t-1} - \frac{1}{t+1000}\nabla w$.\;
        }
\end{algorithm2e}

\subsection{Learning a reference point}\label{sec:learning_ref}
\begin{figure}[!t]
  \centering
  \includegraphics[trim={4cm 5.5cm 4.5cm 5.5cm},clip,width=0.9\linewidth]{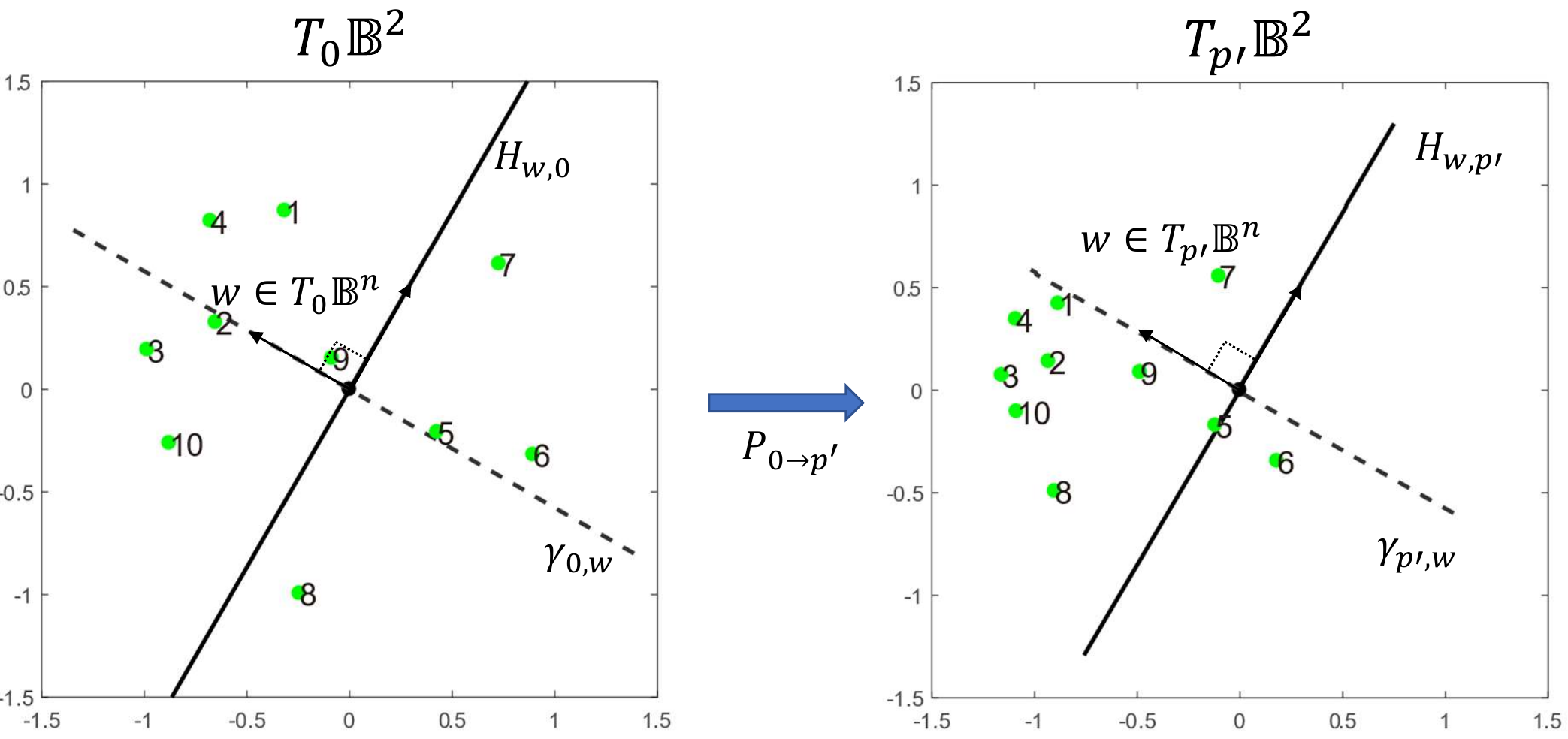}
  \caption{The effect of changing the reference point via parallel transport in corresponding tangent spaces $T_0\mathbb{B}^2$ and $T_{p'}\mathbb{B}^2$, for $p'\in\mathbb{B}^2$. Note that the images of data points, $\log_p(x_i)$, change in the tangent spaces with the choice of the reference point.}\label{fig:2}
\end{figure}

So far we have tacitly assumes that the reference point $p$ is known in advance. While the reference point and normal vector can be learned in a simple manner in Euclidean spaces, this is not the case for the Poincar\'e ball model due to the non-linearity of its logarithmic map and M\"obius addition (Figure~\ref{fig:2}). 
\begin{figure}[!t]
  \centering
    \includegraphics[trim={3cm 4.5cm 3cm 4.5cm},clip,width=0.7\linewidth]{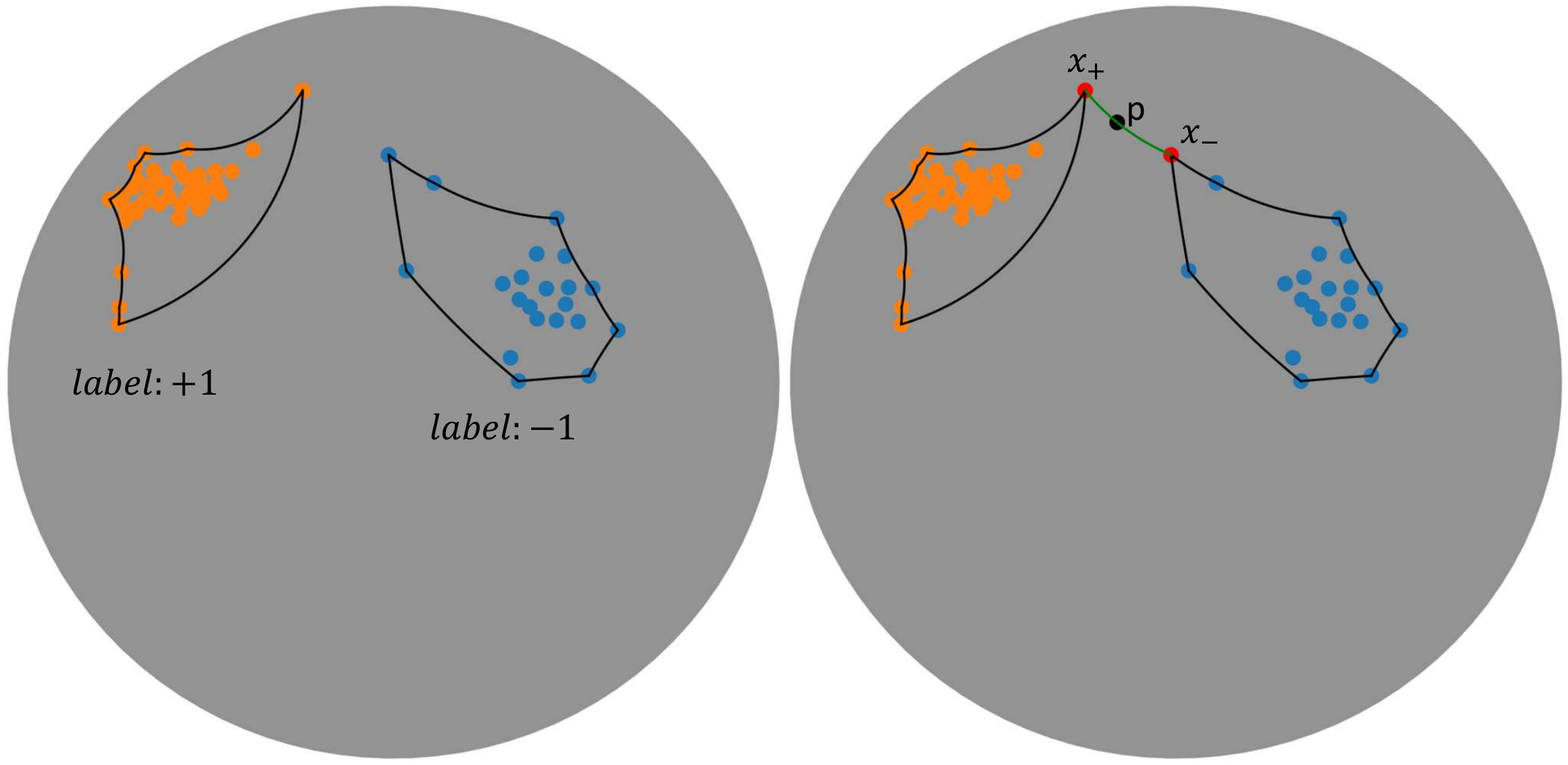}
  \caption{Learning a reference point $p$. Step 1 (left): Construct convex hull for each cluster. Black lines are geodesics defining the surface of convex hull. Step 2 (right): Find a minimum distance pair and choose $p$ as their midpoint.}\label{fig:3}
\end{figure}

Importantly, we have the following observation: A hyperplane correctly classifies all points if and only if it separates their convex hulls (the definition of ``convexity'' in hyperbolic spaces follows from replacing lines with geodesics~\cite{ratcliffe2006foundations}). Hence we can easily generalize known convex hull algorithms to the Poincar\'e ball model, including the Graham scan~\cite{graham1972efficient} and Quickhull~\cite{barber1996quickhull} (see the Appendix for further discussion). Note that the described convex hull algorithm has complexity 
$O(N\log N)$ and is hence very efficient. Next, denote the set of points on the convex hull of the class labeled by $+1$ ($-1$) as $CH_{+}$ ($CH_{-}$). A minimum distance pair $x_+,x_-$ can be found as
\begin{align}
    d(x_+,x_-) = \min_{x\in CH_+,y\in CH_-} d(x,y).
\end{align}
Then, the hyperbolic midpoint of $x_+,x_-$ corresponds to the reference point $p = \gamma_{x_+\rightarrow x_-}(0.5)$ (see Figure~\ref{fig:3}). Our strategy of learning $p$ along with algorithms introduced in Section~\ref{sec:GA2P} works well on real world data set, see Section~\ref{sec:experiments}.

\section{Experiments}\label{sec:experiments}

To put the performance of our proposed algorithms in the context of existing works on hyperbolic classification, we perform extensive numerical experiments on both synthetic and real-world data sets. In particular, we compare our Poincar\'e perceptron, second-order perceptron and SVM method with the hyperboloid SVM of~\cite{cho2019large} and the Euclidean SVM. Detailed descriptions of the experimental settings are provided in the Appendix.
\begin{figure}[!t]
    \centering
    \includegraphics[trim={7.5cm 6cm 5.5cm 6cm},clip,width=0.8\linewidth]{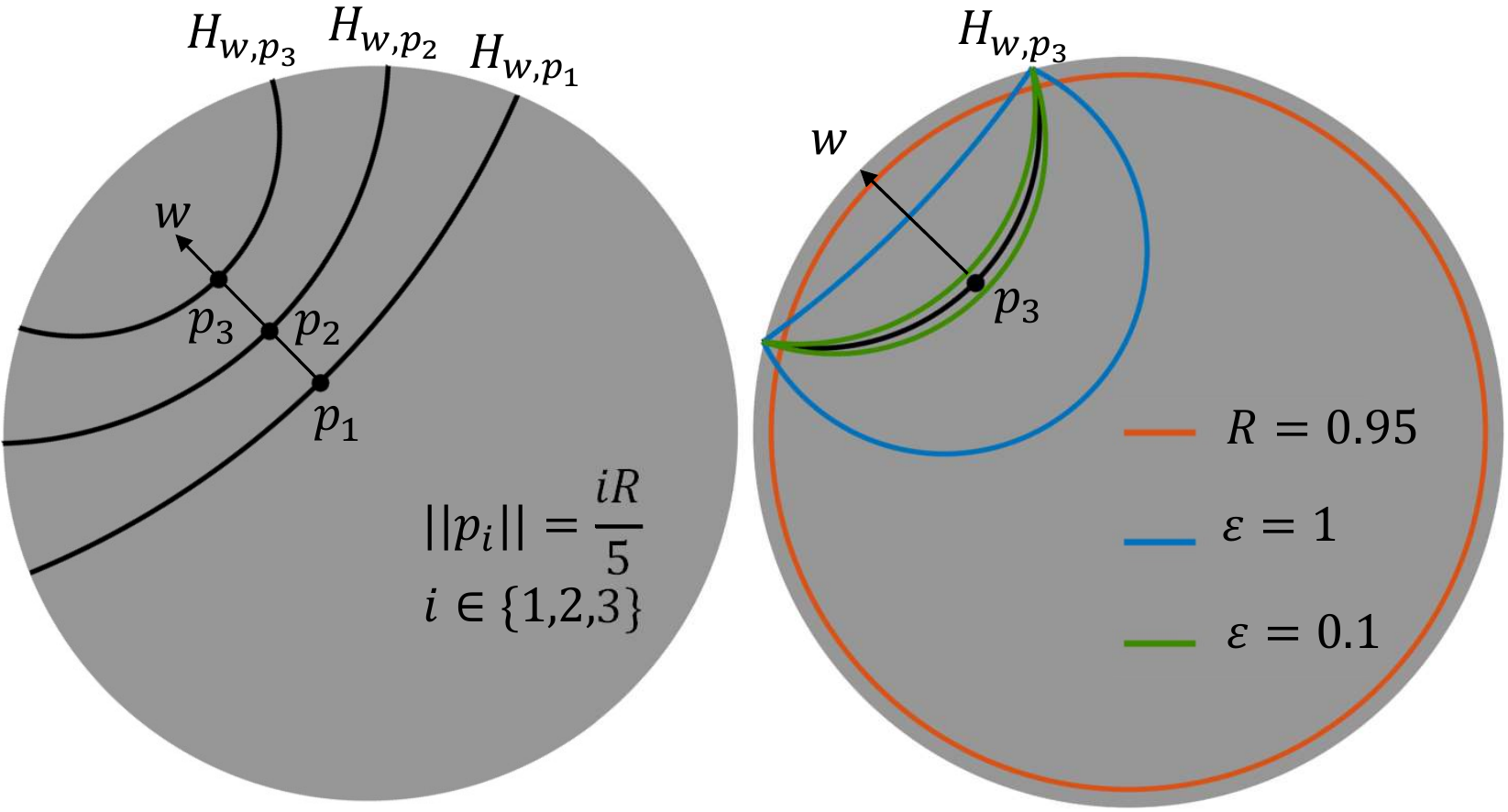}
    \caption{(left) Decision boundaries for different choices of $p$. (right) Geometry of different choices of margin $\varepsilon$.}
    \label{fig:5}
\end{figure}
\subsection{Synthetic data sets}
\begin{figure*}[!t]
  \centering
  \subfigure[Accuracy vs $d$, $\|p\|=\frac{R}{5}$]{\includegraphics[trim={0cm 0cm 0cm 0cm},clip,width=0.24\linewidth]{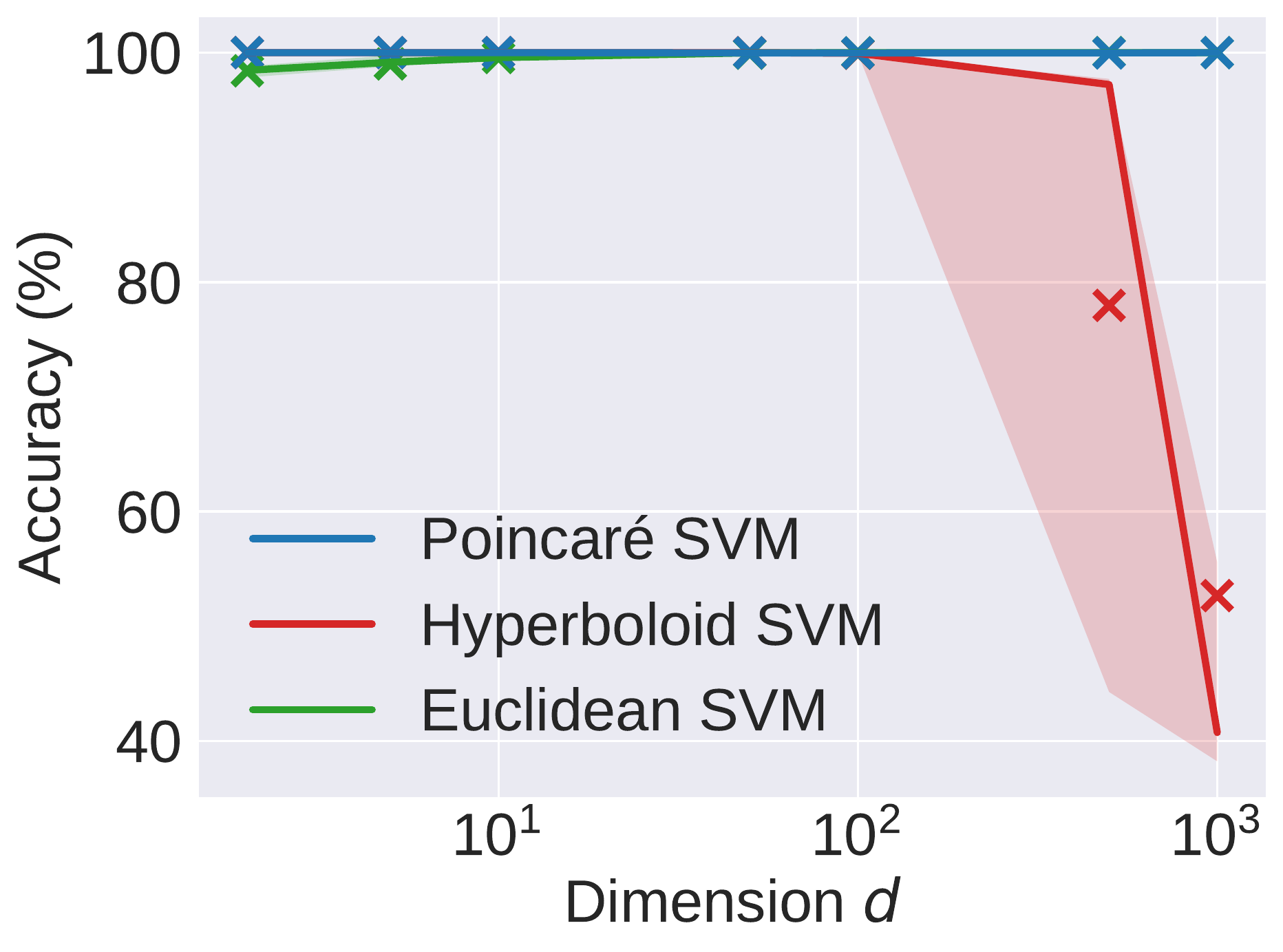}}
  \subfigure[Accuracy vs $d$, $\|p\|=\frac{3R}{5}$]{\includegraphics[trim={0cm 0cm 0cm 0cm},clip,width=0.24\linewidth]{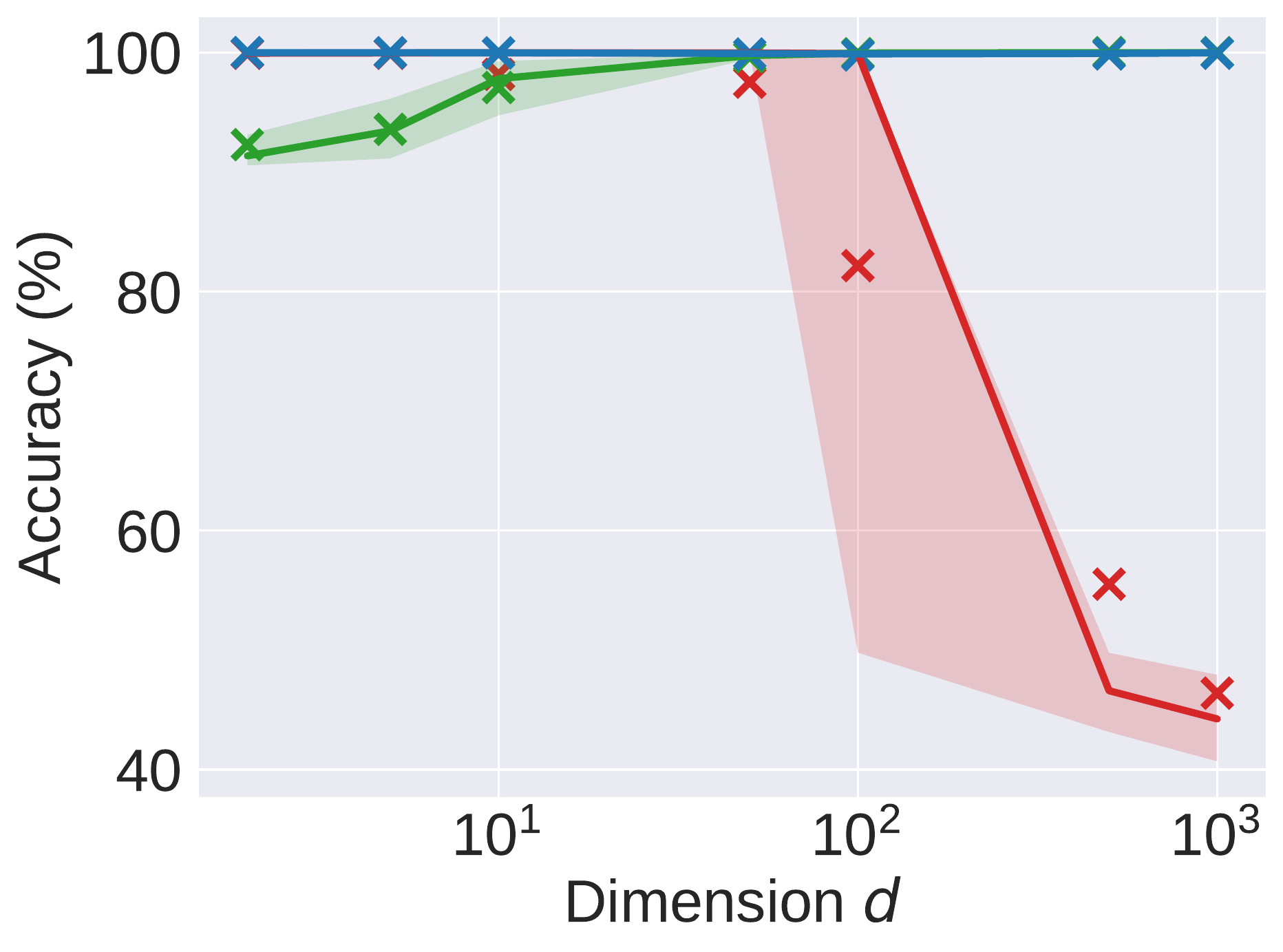}}
  \subfigure[Time vs $d$, $\|p\|=\frac{R}{5}$]{\includegraphics[trim={0cm 0cm 0cm 0cm},clip,width=0.24\linewidth]{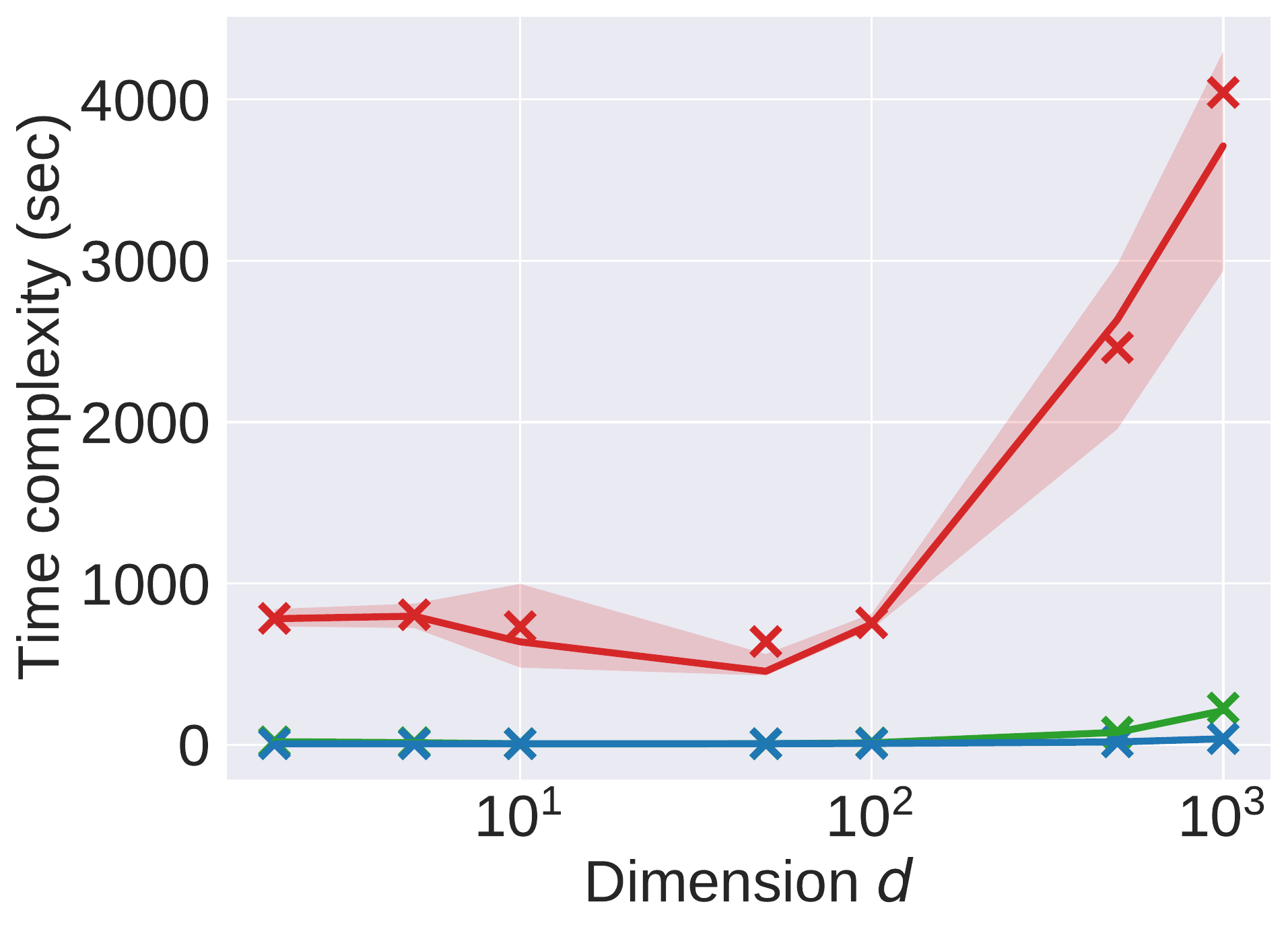}}
  \subfigure[Time vs $d$, $\|p\|=\frac{3R}{5}$]{\includegraphics[trim={0cm 0cm 0cm 0cm},clip,width=0.24\linewidth]{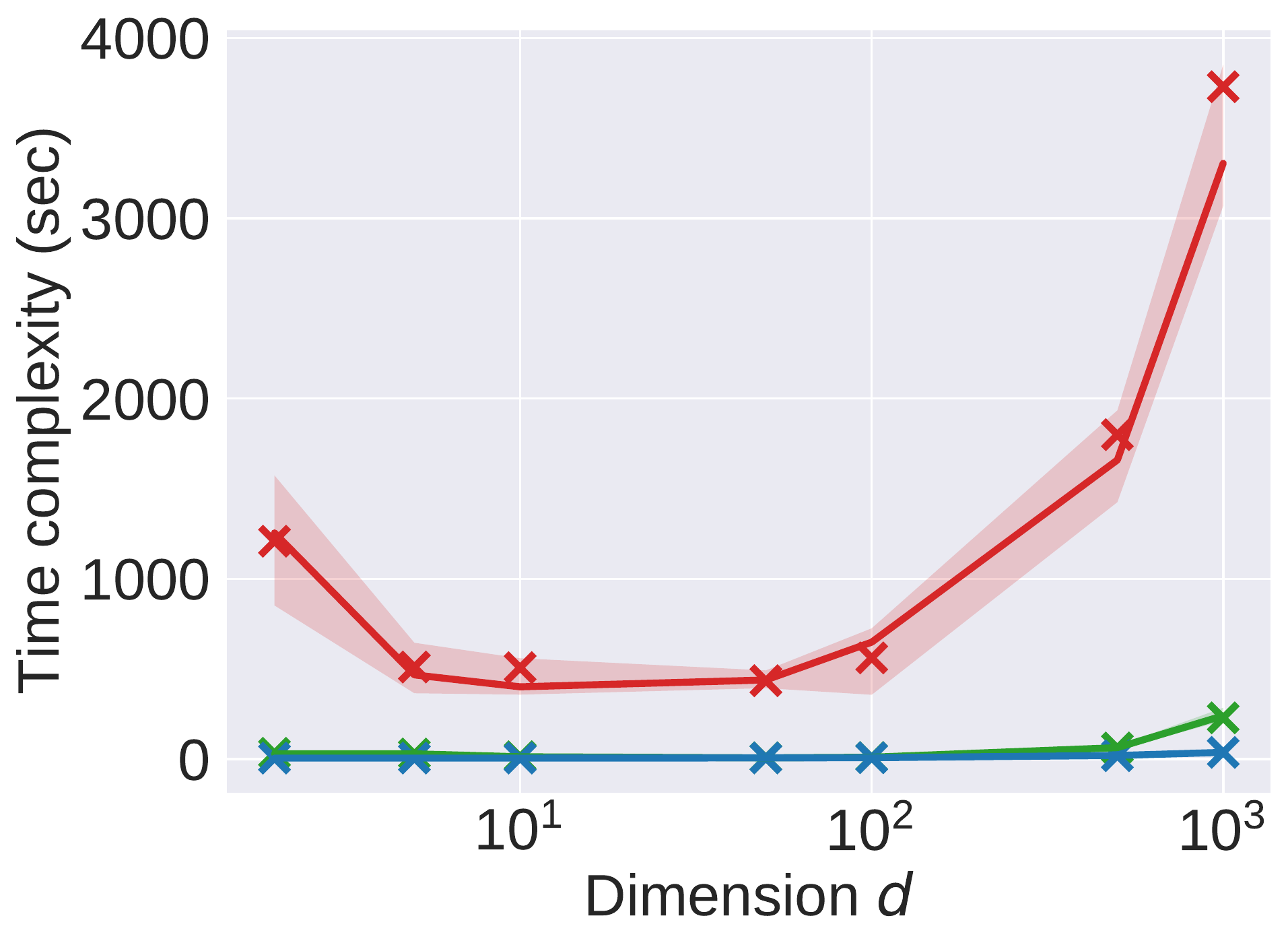}}\\
  \subfigure[Accuracy vs $N$, $\|p\|=\frac{R}{5}$]{\includegraphics[trim={0cm 0cm 0cm 0cm},clip,width=0.24\linewidth]{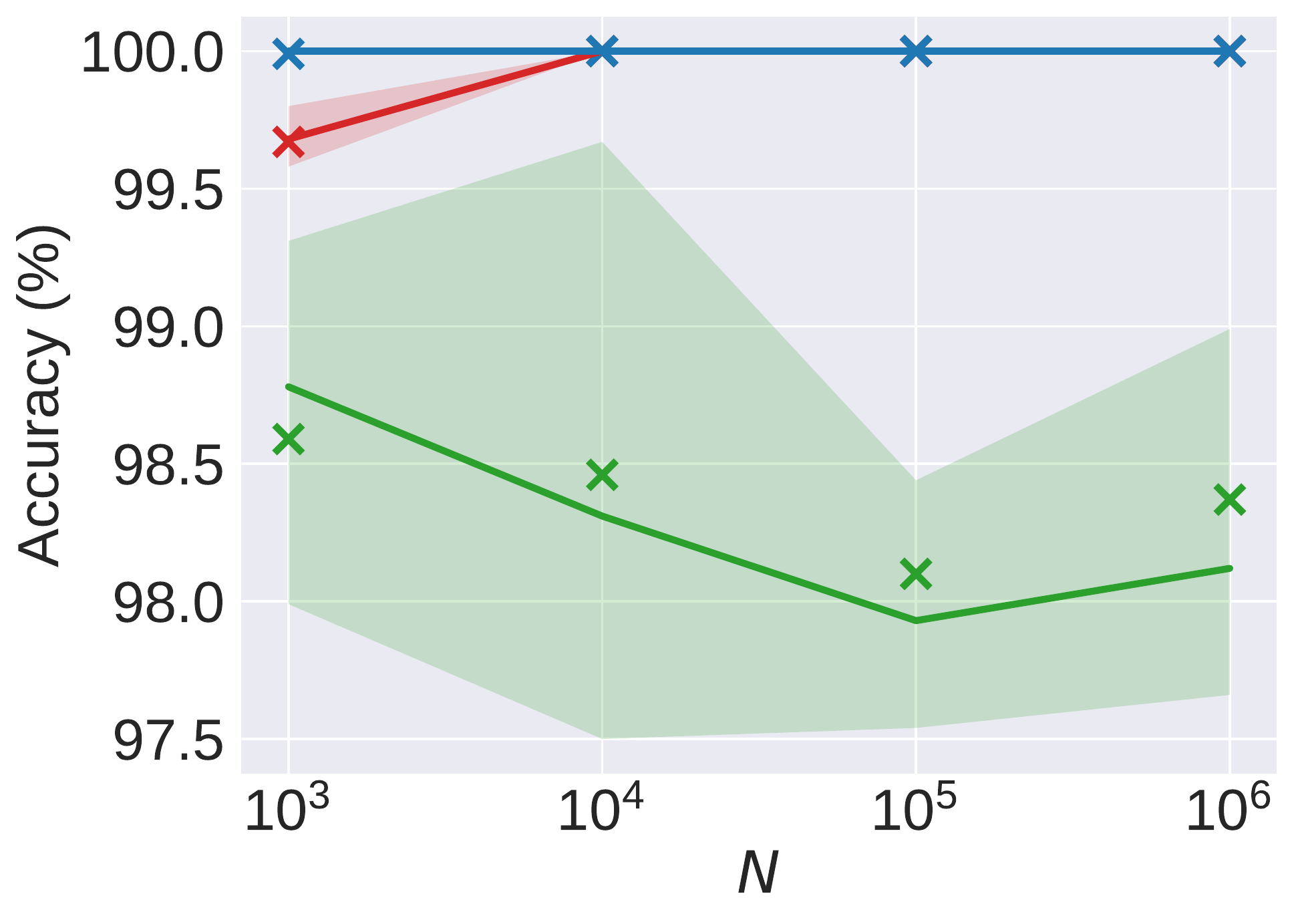}}
  \subfigure[Accuracy vs $N$, $\|p\|=\frac{3R}{5}$]{\includegraphics[trim={0cm 0cm 0cm 0cm},clip,width=0.24\linewidth]{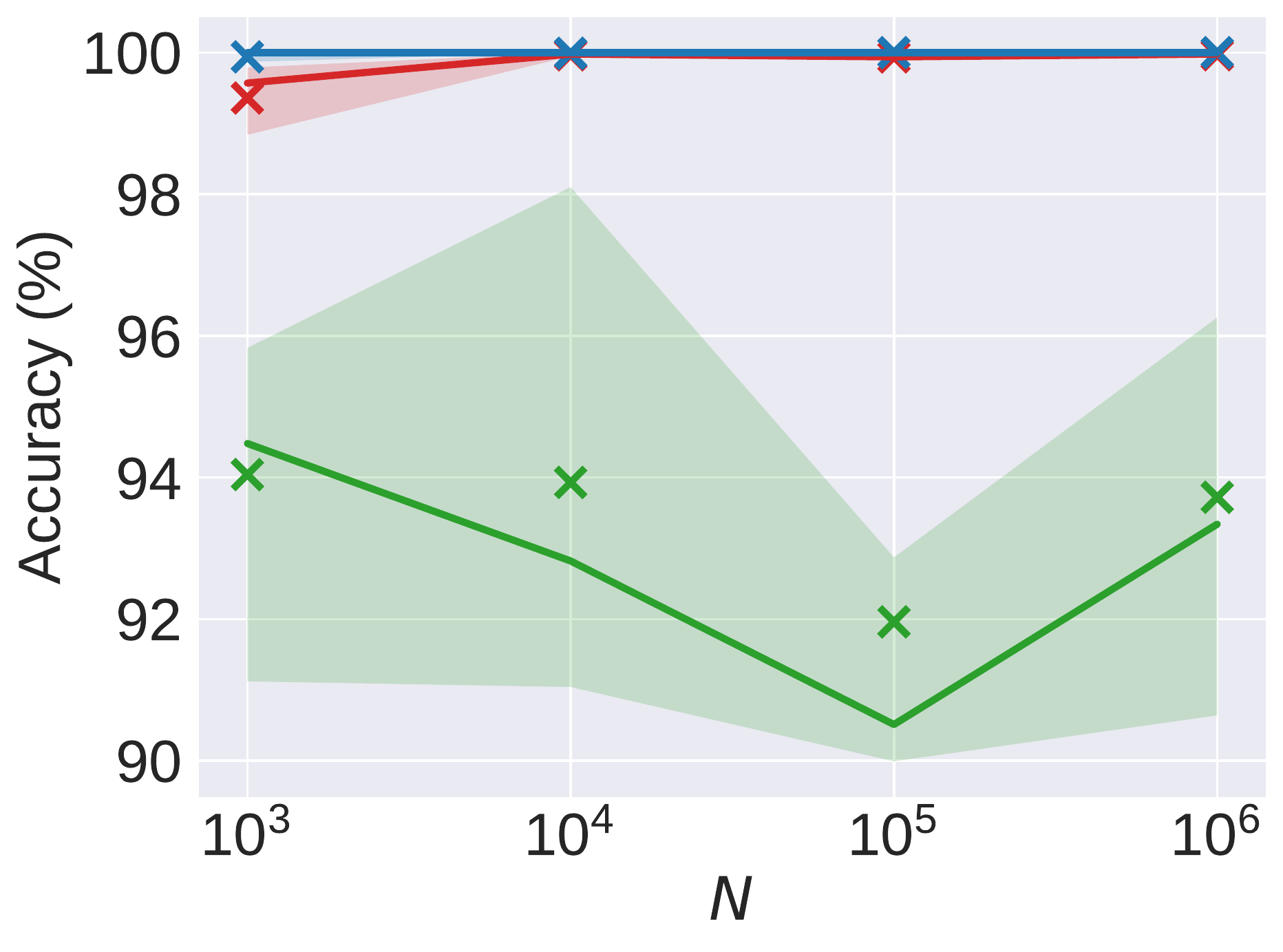}}
  \subfigure[Time vs $N$, $\|p\|=\frac{R}{5}$]{\includegraphics[trim={0cm 0cm 0cm 0cm},clip,width=0.24\linewidth]{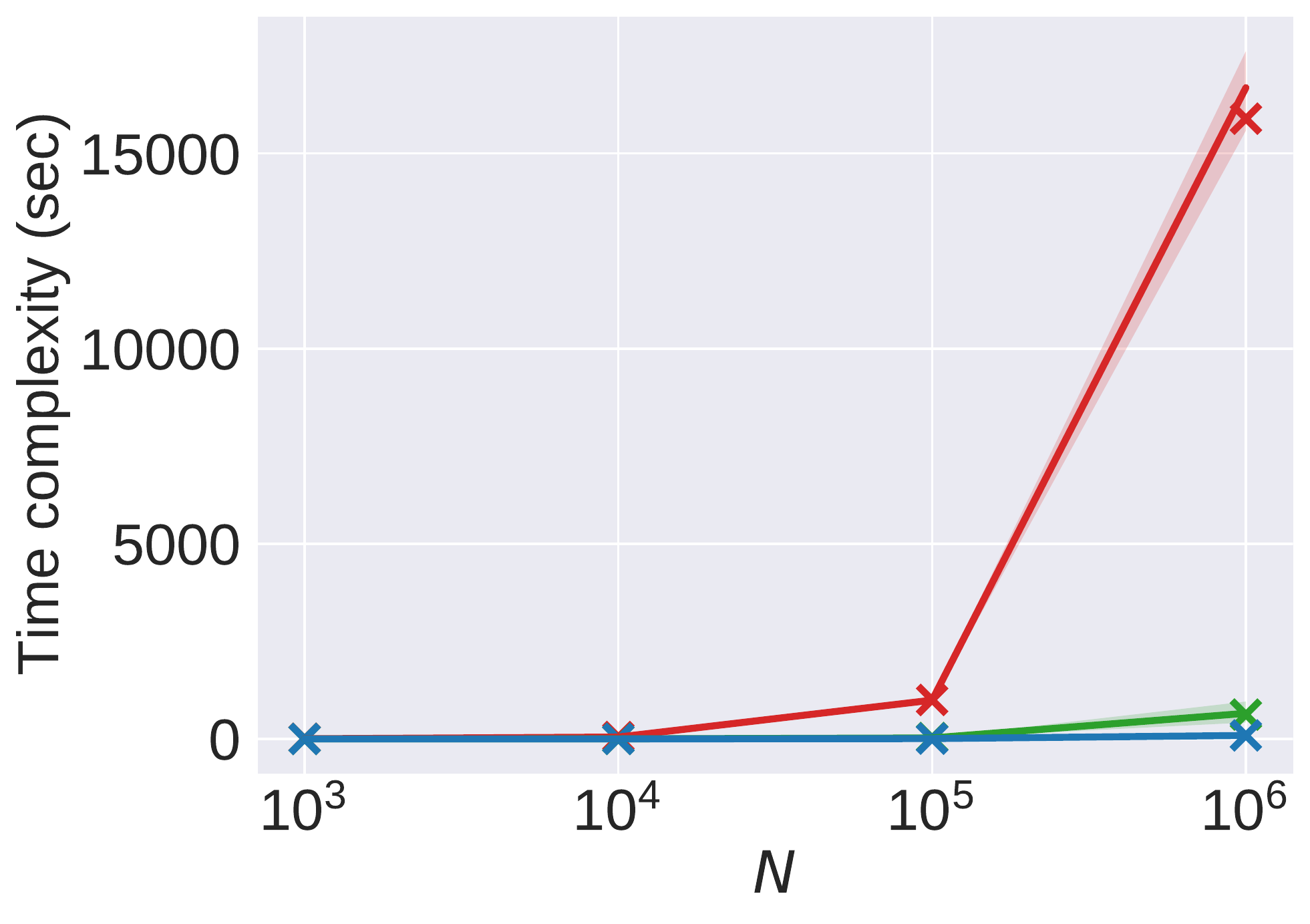}}
  \subfigure[Time vs $N$, $\|p\|=\frac{3R}{5}$]{\includegraphics[trim={0cm 0cm 0cm 0cm},clip,width=0.24\linewidth]{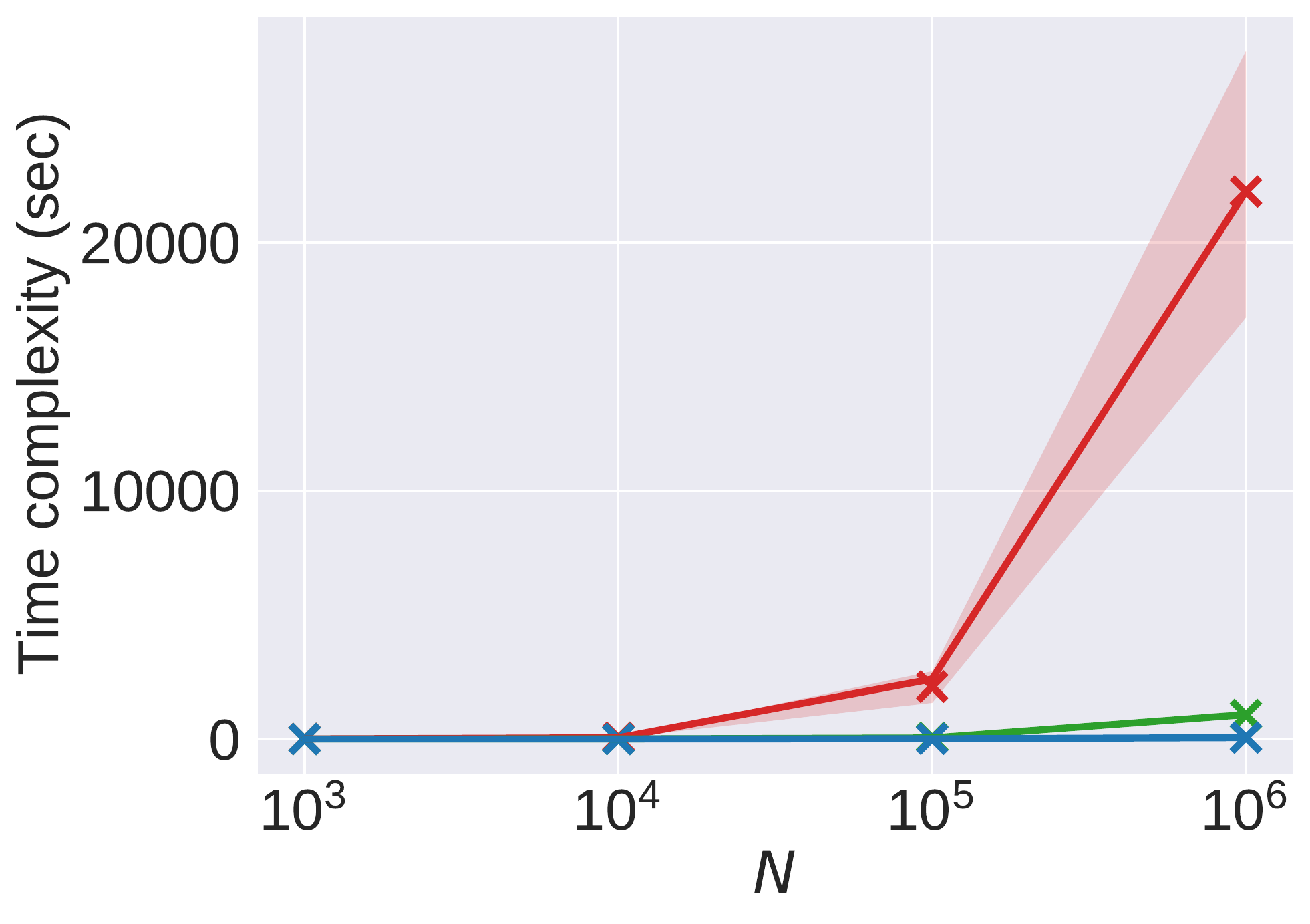}}\\
  \subfigure[Accuracy vs Margin $\varepsilon$, $\|p\|=\frac{R}{5}$]{\includegraphics[trim={0cm 0cm 0cm 0cm},clip,width=0.24\linewidth]{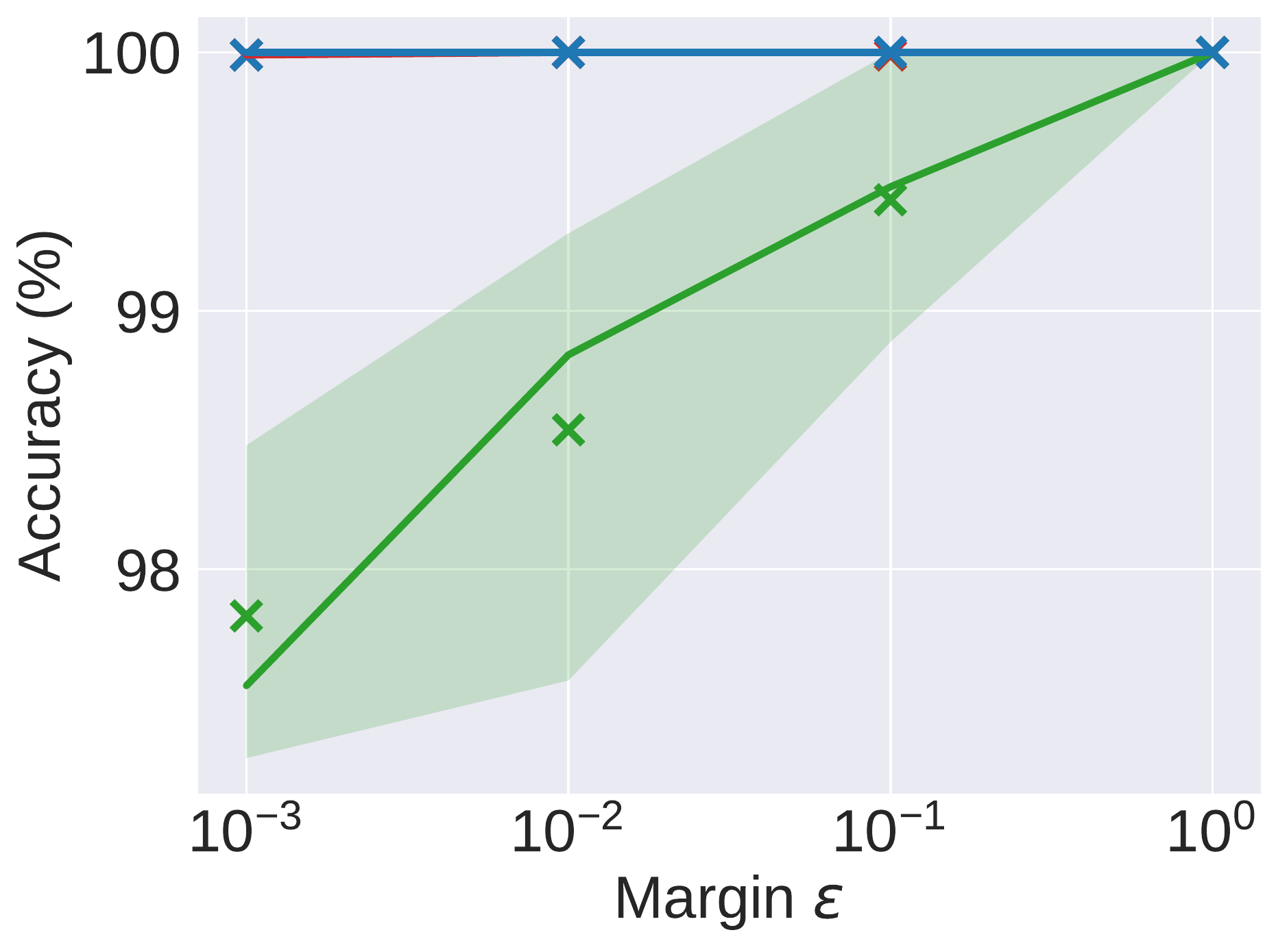}}
  \subfigure[Accuracy vs Margin $\varepsilon$, $\|p\|=\frac{3R}{5}$]{\includegraphics[trim={0cm 0cm 0cm 0cm},clip,width=0.24\linewidth]{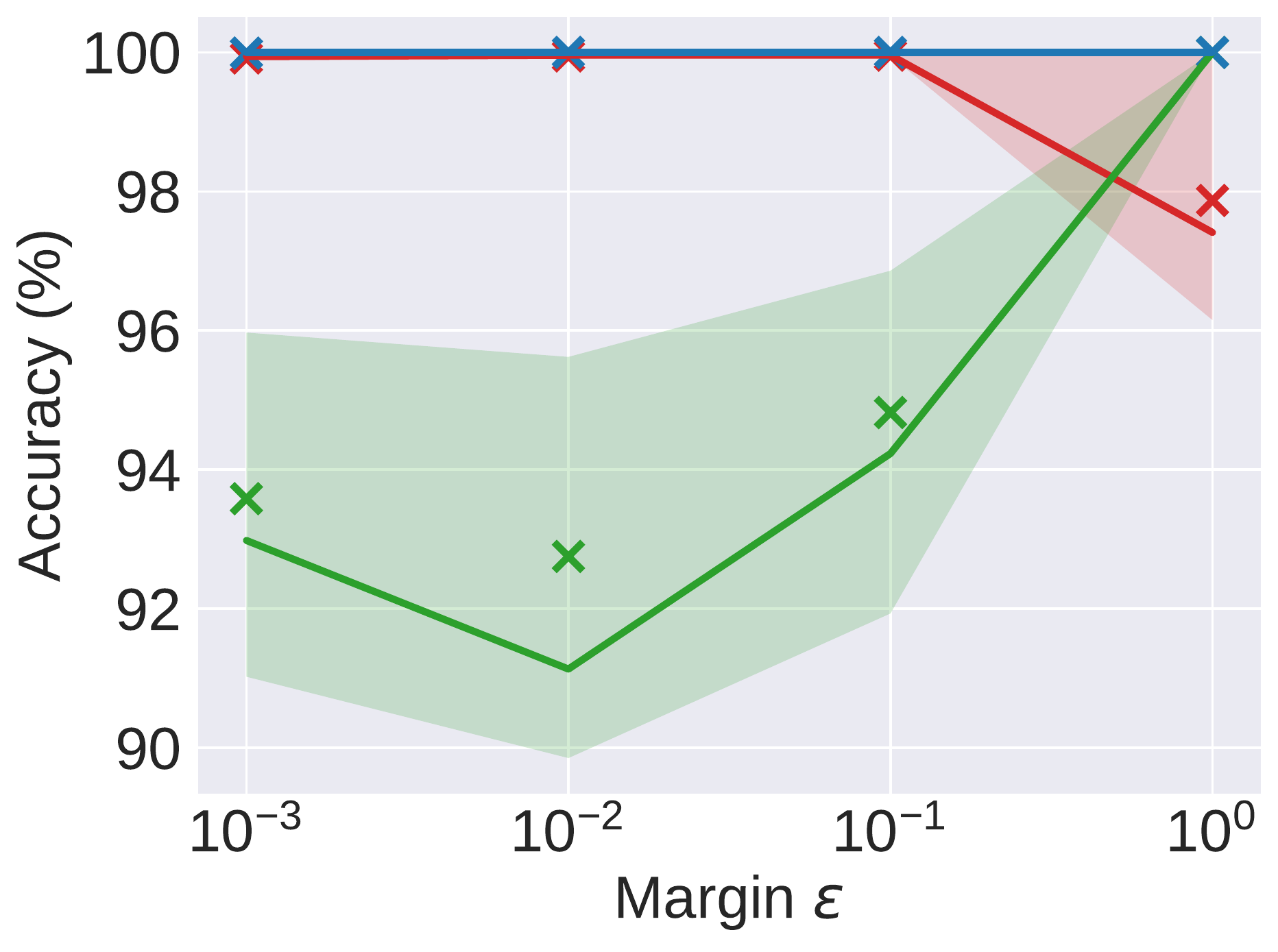}}
  \subfigure[Time vs Margin $\varepsilon$, $\|p\|=\frac{R}{5}$]{\includegraphics[trim={0cm 0cm 0cm 0cm},clip,width=0.24\linewidth]{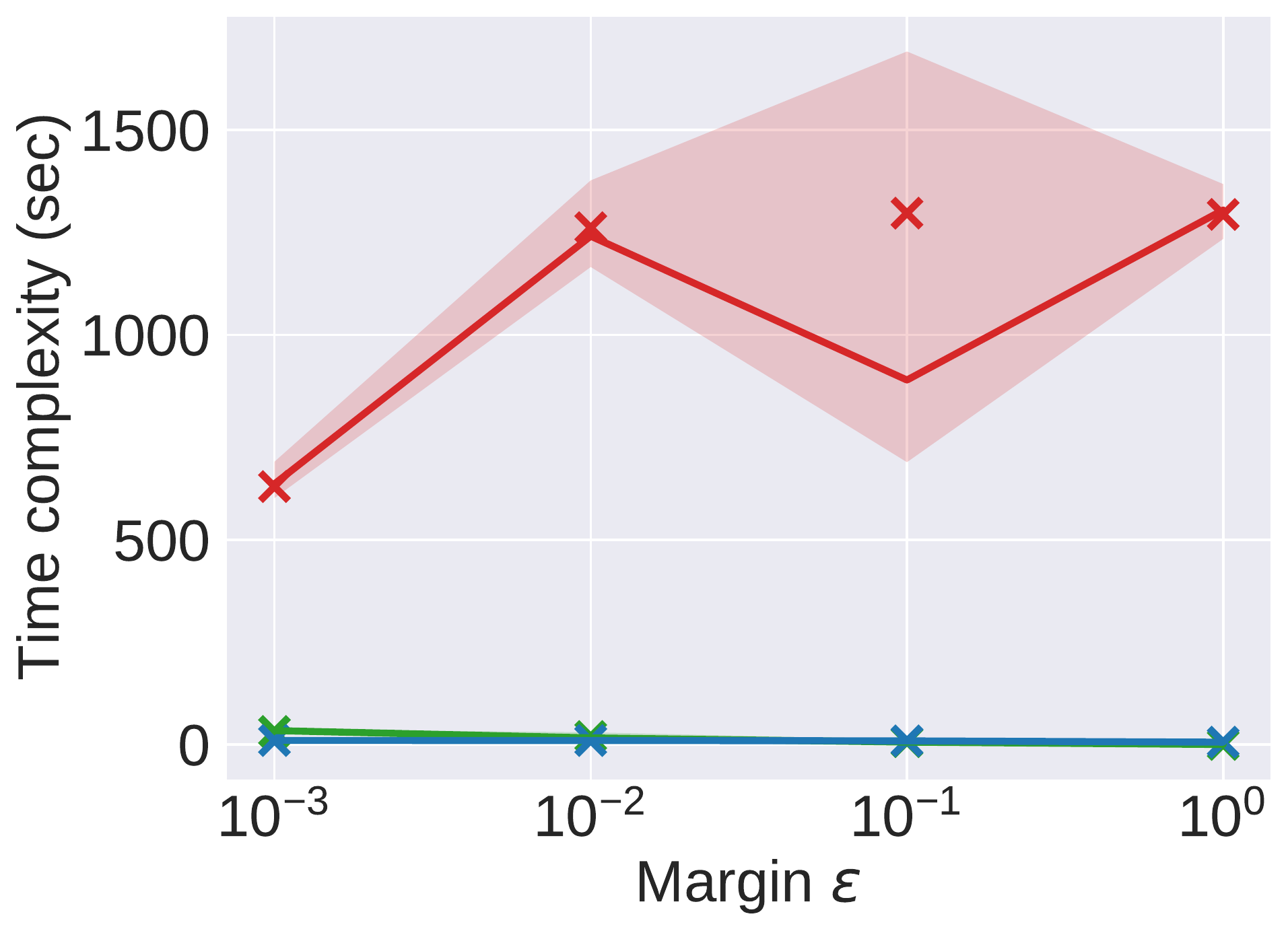}}
  \subfigure[Time vs Margin $\varepsilon$, $\|p\|=\frac{3R}{5}$]{\includegraphics[trim={0cm 0cm 0cm 0cm},clip,width=0.24\linewidth]{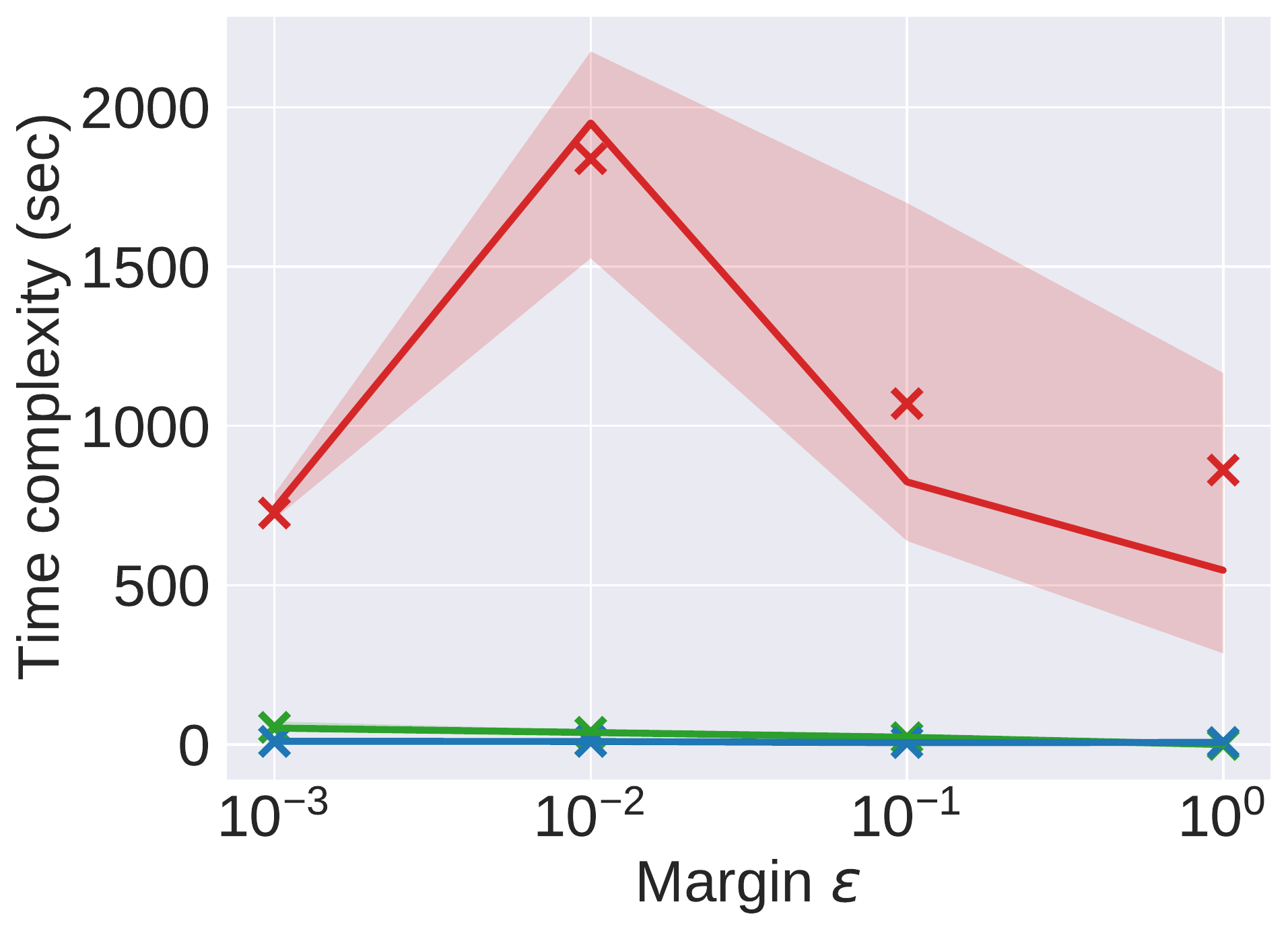}}\\
  \caption{Experiments on synthetic data sets and $\|p\| \in \{{\frac{R}{5},\frac{3R}{5}\}}$. The upper and lower boundaries of the shaded region represent the first and third quantile, respectively. The line itself corresponds to the medium (second quantile) and the marker $\times$ indicates the mean. The first two columns plot the accuracy of the SVM methods while the last two columns plot the corresponding time complexity. For the first row we vary the dimension $d$ from $2$ to $1,000$. For the second row we vary the number of points $N$ from $10^3$ to $10^6$. In the third row we vary the margin $\varepsilon$ from $1$ to $0.001$. The default setting for $(d,N,\varepsilon)$ is $(2,10^5,0.01)$.}\label{fig:4}
\end{figure*}

In the first set of experiments, we generate $N$ points uniformly at random on the Poincar\'e disk and perform binary classification task. To satisfy our Assumption~\ref{asp:all}, we restrict the points to have norm at most $R = 0.95$ (boundedness condition). For a decision boundary $H_{w,p}$, we remove all points within margin $\varepsilon$ (margin assumption). Note that the decision boundary looks more ``curved'' when $\|p\|$ is larger, which makes it more different from the Euclidean case (Figure~\ref{fig:5}). When $\|p\|=0$ then the optimal decision boundary is also linear in Euclidean sense. On the other hand, if we choose $\|p\|$ too large then it is likely that all points are assigned with the same label. Hence, we consider the case $\|p\| = \frac{R}{5},\frac{2R}{5}$ and $\frac{3R}{5}$ and let the direction of $w$ to be generated uniformly at random. Results for case $\|p\|=\frac{2R}{5}$ are demonstrated in Figure~\ref{fig:motivation} while the others are in Figure~\ref{fig:4}. All results are averaged over $20$ independent runs.

We first vary $N$ from $10^3$ to $10^6$ and fix $(d,\varepsilon) = (2,0.01)$. The accuracy and time complexity are shown in Figure~\ref{fig:4} (e)-(h). One can clearly observe that the Euclidean SVM fails to achieve a $100\%$ accuracy as data points are not linearly separable in the Euclidean, but only in hyperbolic sense. This phenomenon becomes even more obvious when $\|p\|$ increases due to the geometry of Poincar\'e disk (Figure~\ref{fig:5}). On the other hand, the hyperboloid SVM is not scalable to accommodate such a large number of points. As an example, it takes \emph{$6$ hours} ($9$ hours for the case $\|p\|=\frac{2R}{5}$, Figure~\ref{fig:motivation}) to process $N=10^6$ points; in comparison, our Poincar\'e SVM only takes \emph{$1$ minute}. Hence, only the Poincar\'e SVM is highly scalable and offers the highest accuracy achievable.

Next we vary the margin $\varepsilon$ from $1$ to $0.001$ and fix $(d,N) = (2,10^5)$. The accuracy and time complexity are shown in Figure~\ref{fig:4} (i)-(l). As the margin reduces, the accuracy of the Euclidean SVM deteriorates. This is again due to the geometry of Poincar\'e disk (Figure~\ref{fig:5}) and the fact that the classifier needs to be tailor-made for hyperbolic spaces. Interestingly, the hyperboloid SVM performs poorly for $\|p\| = 3R/5$ and a margin value $\varepsilon=1$, as its accuracy is significantly below $100\%$. This may be attributed to the fact that the cluster sizes are highly unbalanced, which causes numerical issue with the underlying optimization process. Once again, the Poincar\'e SVM outperforms all other methods in terms of accuracy and time complexity.

Finally, we examined the influence of the data point dimension on the performance of the classifiers. To this end, we varied the dimension $d$ from $2$ to $1,000$ and fixed $(N,\varepsilon) = (10^5,0.01)$. The accuracy and time complexity results are shown in Figure~\ref{fig:4} (a)-(d). Surprisingly, the hyperboloid SVM fails to learn well when $d$ is large and close to $1,000$. This again reaffirms the importance of the convex formulation of our Poincar\'e SVM, which is guaranteed to converge to a global optimum independent of the choice $d,N$ and $\varepsilon$. We also find that Euclidean SVM improves its performance as $d$ increases, albeit at the price of high execution time.

We now we turn our attention to the evaluation of our perceptron algorithms which are online algorithms in nature. Results are summarized in Table~\ref{tab:table1}. There, one can observe that the Poincar\'e second-order perceptron requires a significantly smaller number of updates compared to Poincar\'e perceptron, especially when the margin is small. This parallels the results observed in Euclidean spaces~\cite{cesa2005second}. Furthermore, we validate our Theorem~\ref{thm:HPbound_p} which provide an upper bound on the number of updates for the worst case.

\begin{table}[!t]
\setlength{\tabcolsep}{4pt}
\centering
\scriptsize
\caption{Averaged number of updates for the Poincar\'e second-order perceptron (S-perceptron) and Poincar\'e perceptron for a varying margin $\varepsilon$ and fixed $(N,d) = (10^4,10)$. Bold numbers indicate the best results, with the maximum number of updates over $20$ runs shown in parenthesis. Also shown is a theoretical upper bound on the number of updates for the Poincar\'e perceptron based on Theorem~\ref{thm:HPbound_p}.}
\label{tab:table1}
\begin{tabular}{@{}cccccc@{}}
\toprule
                                      & Margin $\varepsilon$                               & 1           & 0.1               & 0.01                & 0.001               \\ \midrule
\multirow{5}{*}{$\|p\|=\frac{R}{5}$}  & \multicolumn{1}{c|}{\multirow{2}{*}{S-perceptron}} & \textbf{26} & \textbf{82}       & \textbf{342}        & \textbf{818}        \\
                                      & \multicolumn{1}{c|}{}                              & (34)        & (124)             & (548)               & (1,505)              \\
                                      & \multicolumn{1}{c|}{\multirow{2}{*}{perceptron}}   & 51          & 1,495              & $1.96\times 10^4$   & $1.34\times 10^5$   \\
                                      & \multicolumn{1}{c|}{}                              & (65)        & (2,495)            & ($3.38\times 10^4$) & ($3.56\times 10^5$) \\
                                      & \multicolumn{1}{c|}{Theorem~\ref{thm:HPbound_p}}   & 594         & 81,749             & $8.2\times 10^6$    & $8.2\times 10^8$    \\ \midrule
\multirow{5}{*}{$\|p\|=\frac{3R}{5}$} & \multicolumn{1}{c|}{\multirow{2}{*}{S-perceptron}} & \textbf{29} & \textbf{101}      & \textbf{340}        & \textbf{545}        \\
                                      & \multicolumn{1}{c|}{}                              & (41)        & (159)             & (748)               & (986)               \\
                                      & \multicolumn{1}{c|}{\multirow{2}{*}{perceptron}}   & 82          & 1,158              & $1.68\times 10^4$   & $1.46\times 10^5$   \\
                                      & \multicolumn{1}{c|}{}                              & (138)       & (2,240)            & ($6.65\times 10^4$) & ($7.68\times 10^5$) \\
                                      & \multicolumn{1}{c|}{Theorem~\ref{thm:HPbound_p}}   & 3,670        & $5.05\times 10^5$ & $5.07\times 10^7$   & $5.07\times 10^9$   \\ \bottomrule
\end{tabular}
\end{table}

\subsection{Real-world data sets}

For real-world data sets, we choose to work with more than two collections of points. To enable $K$-class classification for $K \geq 2$, we use $K$ binary classifiers that are independently trained on the same training set to separate each single class from the remaining classes. For each classifier, we transform the resulting prediction scores into probabilities via the Platt scaling technique~\cite{platt1999probabilistic}. The predicted labels are then decided by a maximum a posteriori criteria based on the probability of each class.

The data sets of interest include Olsson's single-cell expression profiles, containing single-cell (sc) RNA-seq expression data from $8$ classes (cell types)~\cite{olsson2016single}, CIFAR10, containing images from common objects falling into $10$ classes~\cite{krizhevsky2009learning}, Fashion-MNIST, containing Zalando's article images with $10$ classes~\cite{xiao2017fashion}, and mini-ImageNet, containing subsamples of images from original ImageNet data set and containing $20$ classes~\cite{ravi2016optimization}. Following the procedure described in~\cite{klimovskaia2020poincare,khrulkov2020hyperbolic}, we embed single-cell, CIFAR10 and Fashion-MNIST data sets into a $2$-dimensional Poincar\'e disk, and mini-ImageNet into a $512$-dimensional Poincar\'e ball, all with curvature $-1$ (Note that our methods can be easily adapted to work with other curvature values as well), see Figure~\ref{fig:real_data}. Other details about the data sets including the number of samples and splitting strategy of training and testing set is described in the Appendix. Since the real-world embedded data sets are not linearly separable we only report classification results for the Poincar\'e SVM method.

We compare the performance of the Poincar\'e SVM, Hyperbolid SVM and Euclidean SVM for soft-margin classification of the above described data points. For the Poincar\'e SVM the reference point $p^{(i)}$ for each binary classifier is estimated via our technique introduced in Section~\ref{sec:learning_ref}. The resulting classification accuracy and time complexity are shown in Table~\ref{tab:real_performance}. From the results one can easily see that our Poincar\'e SVM consistently achieves the best classification accuracy over all data sets while being roughly $10$x faster than the hyperboloid SVM. It is also worth pointing out that for most data sets embedded into the Poincar\'e ball model, the Euclidean SVM method does not perform well as it does not exploit the geometry of data; however, the good performance of the Euclidean SVM algorithm on mini-ImageNet can be attributed to the implicit Euclidean metric used in the embedding framework of~\cite{khrulkov2020hyperbolic}. Note that since the Poincar\'e SVM and Euclidean SVM are guaranteed to achieve a global optimum, the standard deviation of classification accuracy is zero.

\begin{table}[!t]
\setlength{\tabcolsep}{4pt}
\centering
\scriptsize
\caption{Performance of the SVM algorithms generated based on $5$ 
independent trials.}
\label{tab:real_performance}
\begin{tabular}{@{}cccc@{}}
\toprule
                                      & Algorithm   & Accuracy (\%)           & Time (sec)               \\ \midrule
\multirow{3}{*}{Olsson's scRNA-seq}  & \multicolumn{1}{c|}{{Euclidean SVM}} & {$71.59\pm 0.00$} & {$0.06\pm 0.00$} \\
                                      & \multicolumn{1}{c|}{{Hyperboloid SVM}}   & $71.97\pm 0.54$          & $4.49\pm0.05$ \\
                                      & \multicolumn{1}{c|}{Poincar\'e SVM}   & {$\mathbf{89.77\pm 0.00}$}         & $0.16\pm 0.00$ \\ \midrule
\multirow{3}{*}{CIFAR10} & \multicolumn{1}{c|}{{Euclidean SVM}} & $47.66\pm 0.00$ & $26.03\pm 5.29$ \\
                                      & \multicolumn{1}{c|}{{Hyperboloid SVM}}   & $89.87\pm 0.01$          & $707.69\pm 15.33$   \\
                                      & \multicolumn{1}{c|}{Poincar\'e SVM}   & $\mathbf{91.84\pm 0.00}$        & $45.55\pm 0.43$ \\ \midrule
\multirow{3}{*}{Fashion-MNIST} & \multicolumn{1}{c|}{{Euclidean SVM}} & $39.91\pm 0.01$ & $32.70\pm 11.11$ \\
                                      & \multicolumn{1}{c|}{{Hyperboloid SVM}}   & $76.78\pm 0.06$          & $898.82\pm 11.35$   \\
                                      & \multicolumn{1}{c|}{Poincar\'e SVM}   & $\mathbf{87.82\pm 0.00}$        & $67.28\pm 8.63$ \\ \midrule
\multirow{3}{*}{mini-ImageNet} & \multicolumn{1}{c|}{{Euclidean SVM}} & $63.33\pm 0.00$ & $7.94\pm 0.09$ \\
                                      & \multicolumn{1}{c|}{{Hyperboloid SVM}}   & $31.75\pm 1.91$          & $618.23\pm 10.92$   \\
                                      & \multicolumn{1}{c|}{Poincar\'e SVM}   & $\mathbf{63.59\pm 0.00}$        & $18.78\pm 0.42$ \\ \bottomrule
\end{tabular}
\end{table}

\section{Conclusion}
We generalize classification algorithms such as (second-order) perceptron and SVM to Poincar\'e balls. Our Poincar\'e classification algorithms comes with theoretical guarantee of converging to global optimum which improves the previous attempts in the literature. We validate our Poincar\'e classification algorithms with experiments on both synthetic and real-world datasets. It shows that our method is highly scalable and accurate, which aligns with our theoretical results. Our methodology appears to be amenable for extensions to other machine learning problems in hyperbolic geometry. One example, pertaining to classification in mixed constant curvature spaces, can be found in~\cite{tabaghi2021linear}.

\section*{Acknowledgment}
The work was supported in part by the NSF grant 1956384.

\bibliography{example_paper}
\bibliographystyle{IEEEtran}

\newpage
\onecolumn
\appendix
\section*{Proof of Lemma~\ref{lma:paramax_mobadd}}
By the definition of M\"obius addition, we have
\begin{align*}
    a\oplus b = \frac{1+2a^{T}b+\|b\|^2}{1+2a^{T}b+\|a\|^2\|b\|^2}a + \frac{1-\|a\|^2}{1+2a^{T}b+\|a\|^2\|b\|^2}b.
\end{align*}
 Thus, 
 \begin{align}
    &\|a\oplus b\|^2 = \frac{\| (1+2a^Tb+\|b\|^2)a + (1-\|a\|^2)b\|^2}{ \big( 1+2a^Tb+\|a\|^2\|b\|^2 \big)^2} \nonumber \\
    &= \frac{\|a+b\|^2 \big(1+\|b\|^2\|a\|^2+2a^Tb \big) }{ \big( 1+2a^Tb+\|a\|^2\|b\|^2 \big)^2} =\frac{\|a+b\|^2}{1+2a^Tb+\|a\|^2\|b\|^2} \label{eq:aoplusbsq2}.
 \end{align}
Next, use $\|b\|=r$ and $a^Tb = r\|a\|\cos(\theta)$ in the above expression: 
 \begin{align}
     \|a\oplus b\|^2  &= \frac{\|a\|^2+2r\|a\|\cos(\theta) + r^2}{1+2r\|x\|\cos(\theta) + \|a\|^2r^2} \nonumber \\
      &=1-\frac{(1-r^2)(1-\|a\|^2)}{1+2r\|x\|\cos(\theta) + \|a\|^2r^2}. \label{eq:aoplusbsq}
 \end{align}
The function in~\eqref{eq:aoplusbsq} attains its maximum at $\theta = 0$ and $r=R$. We also observe that~\eqref{eq:aoplusbsq2} is symmetric in $a,b$. Thus, the same argument holds for $\|b\oplus a\|$. 

\section*{Proof of Theorem~\ref{thm:PSOP}}
We generalize the arguments in~\cite{cesa2005second} to hyperbolic spaces. Let $A_0 = aI$.  The matrix $A_k$ can be recursively computed from $A_k = A_{k-1} + z_tz_t^T$, or equivalently $A_k = aI + X_kX_k^T$. Without loss of generality, let $t_k$ be the time index of the $k^{th}$ error. 
 \begin{align*}
    & \xi_k^T A_k^{-1} \xi_k = (\xi_{k-1} + y_{t_k}z_{t_k})^TA_k^{-1}(\xi_{k-1} + y_{t_k}z_{t_k}) \\
    & = \xi_{k-1}^TA_k^{-1}\xi_{k-1} + z_{t_k}^TA_k^{-1}z_{t_k} + 2y_{t_k}(A_k^{-1}\xi_{k-1})^Tz_{t_k}\\
    & = \xi_{k-1}^TA_k^{-1}\xi_{k-1} + z_{t_k}^TA_k^{-1}z_{t_k} + 2y_{t_k}(w_{t_k})^Tz_{t_k}\\
    & \leq \xi_{k-1}^TA_k^{-1}\xi_{k-1} + z_{t_k}^TA_k^{-1}z_{t_k}  \\
    & \stackrel{\mathrm{(a)}}{=} \xi_{k-1}^TA_{k-1}^{-1}\xi_{k-1} - \frac{(\xi_{k-1}^TA_{k-1}^{-1}z_{t_k})^2}{1+z_{t_k}^TA_{k-1}^{-1}z_{t_k}} + z_{t_k}^TA_k^{-1}z_{t_k} \\
    & \leq \xi_{k-1}^TA_{k-1}^{-1}\xi_{k-1}+ z_{t_k}^TA_k^{-1}z_{t_k}
 \end{align*}
where $\mathrm{(a)}$ is due to the Sherman-Morrison formula~\cite{sherman1950adjustment} below. 
  \begin{lemma}[\cite{sherman1950adjustment}]\label{lma:SMformula}
  Let $A$ be an arbitrary $n\times n$ positive-definite matrix. Let $x\in \mathbb{R}^n$. Then $B = A+xx^T$ is also a positive-definite matrix and
  \begin{align}
      & B^{-1} = A^{-1} - \frac{(A^{-1}x)(A^{-1}x)^T}{1+x^TA^{-1}x}.
  \end{align}
 \end{lemma}
 Note that the inequality holds since $A_{k-1}$ is a positive-definite matrix and thus so is its inverse. Therefore, we have
 \begin{align}
     &\xi_k^T A_k^{-1} \xi_k \leq \xi_{k-1}^TA_{k-1}^{-1}\xi_{k-1} + z_{t_k}^TA_k^{-1}z_{t_k} \leq \sum_{j\in [k]}z_{t_j}^TA_j^{-1}z_{t_j}\nonumber\\
     & \stackrel{\mathrm{(b)}}{=} \sum_{j\in [k]}\left(1-\frac{\text{det}(A_{j-1})}{\text{det}(A_{j})}\right)\stackrel{\mathrm{(c)}}{\leq} \sum_{j\in [k]}\log(\frac{\text{det}(A_{j})}{\text{det}(A_{j-1})})\nonumber\\
     & = \log(\frac{\text{det}(A_{k})}{\text{det}(A_{0})}) = \log(\frac{\text{det}(aI + X_kX_k^T)}{\text{det}(aI)})= \sum_{i\in[n]}\log(1+\frac{\lambda_i}{a}),\nonumber
 \end{align}
 where $\lambda_i$ are the eigenvalues of $X_kX_k^T$. Claim $\mathrm{(b)}$ follows from Lemma~\ref{lma:det} while $\mathrm{(c)}$ is due to the fact $1-x\leq -\log(x),\forall x>0$.
 \begin{lemma}[\cite{cesa2005second}]\label{lma:det}
  Let $A$ be an arbitrary $n\times n$ positive-semidefinite matrix. Let $x\in \mathbb{R}^n$ and $B = A-xx^T$. Then
  \begin{align}
      x^TA^\dagger x =
      \begin{cases}
        1 & \text{ if } x\notin span(B)\\
        1-\frac{\text{det}_{\neq 0}(B)}{\text{det}_{\neq 0}(A)} <1 & \text{ if } x \in span(B)
      \end{cases},
  \end{align}
  where $\text{det}_{\neq 0}(B)$ is the product of non-zero eigenvalues of $B$.
 \end{lemma}
This leads to the upper bound for $\xi_k^TA_k^{-1}\xi_k$. For the lower bound, we have
 \begin{align}
     &\sqrt{\xi_k^TA_k^{-1}\xi_k} \geq \ip{A_k^{-1/2}\xi_k}{\frac{A_k^{1/2}w^\star}{\|A_k^{1/2}w^\star\|}} = \frac{\ip{\xi_k}{w^\star}}{\|A_k^{1/2}w^\star\|}\geq \frac{k \varepsilon'}{\|A_k^{1/2}w^\star\|}. \nonumber
 \end{align}
 Also, recall $\xi_k = \sum_{j\in[k]}y_{\sigma(j)}z_{\sigma(j)}$. Combining the bounds we get
 \begin{align*}
     (\frac{k \varepsilon'}{\|A_k^{1/2}w^\star\|})^2 \leq \xi_k^TA_k^{-1}\xi_k \leq \sum_{i\in[n]}\log(1+\frac{\lambda_i}{a}).
 \end{align*}
 This leads to the bound $k \leq \frac{\|A_k^{1/2}w^\star\|}{\varepsilon'}\sqrt{\sum_{i\in[n]}\log(1+\frac{\lambda_i}{a})}$. Finally, since $\|w^\star\| = 1$, we have
 \begin{align}
     & \|A_k^{1/2}w^\star\|^2 = (w^\star)^T (aI + X_kX_k^T) w^\star = a + \lambda_{w^\star},\nonumber
 \end{align}
which follows from the definition of $\lambda_{w^\star}$. Hence, 
 \begin{align}
     k \leq \frac{1}{\varepsilon'}\sqrt{(a + \lambda_{w^\star})\sum_{i\in[n]}\log(1+\frac{\lambda_i}{a})},
 \end{align}
which completes the proof.

\section*{Convex hull algorithms in Poincar\'e ball model}
We introduce next a generalization of the Graham scan and Quickhull algorithms for the Poincar\'e ball model. In a nutshell, we replace lines with geodesics and vectors $\overrightarrow{AB}$ with tangent vectors $\log_A(B)$ or equivalently $(-A)\oplus B$. The pseudo code for the Poincar\'e version of the Graham scan is listed in Algorithm~\ref{alg:Graham_scan}, while Quickhull is listed in Algorithm~\ref{alg:Quickhull} (both for the two-dimensional case). The Graham scan has worst-case time complexity $O(N\log N)$, while Quickhull has complexity $O(N\log N)$ in expectation and $O(N^2)$ in the worst-case. The Graham scan only works for two-dimensional points while Quickhull can be generalized for higher dimensions~\cite{barber1996quickhull}.

\begin{algorithm2e}
\caption{Poincar\'e Graham scan}\label{alg:Graham_scan}
\SetAlgoLined
\DontPrintSemicolon
\SetKwInput{Input}{Input}
\SetKwInOut{Output}{Output}
  \Input{Data points $X = \{x_i\}_{i=1}^N\in \mathbb{B}^2$.}
  Initialization: Set $S=\emptyset$.\;
  Find $p_0$ with minimum $y$ coordinate. If multiple options exist, choose the smallest $x$-coordinate one.\;
  In the $T_{p_0}\mathbb{B}^2$, sort $X$ by the angle between $\log_{p_0}(x_i)$ and the $x$-axis (counterclockwise, ascending).\;
  Append an additional $p_0$ to the end of $X$.\;
  \For {$x\in X$}{
        \While{ $|S|>1$ and outer-product$\left(\log_{S[-2]}(S[-1]),\log_{S[-2]}(x)\right)<0$}{
            Pop $S$;\;
        }
        Push $x$ to $S$;\;
        }
  \Output {$S$.}
\end{algorithm2e}

\begin{algorithm2e}
\caption{Poincar\'e Quickhull}\label{alg:Quickhull}
\SetAlgoLined
\DontPrintSemicolon
\SetKwInput{Input}{Input}
\SetKwInOut{Output}{Output}
  \Input{Data points $X = \{x_i\}_{i=1}^N\in \mathbb{B}^2$.}
  Initialization: Set $S=\emptyset$.\;
  Find the left- and right-most points $A,B$. Add them to $S$.\;
  \tcp{$\gamma_{A\rightarrow B}$ splits the remaining points into two groups, $S_1$ and $S_2$.}
  $p_0=\gamma_{A\rightarrow B}(\frac{1}{2})$, $v=\log_{p_0}(B)$, $w = v^\perp$;\;
  \For {$x\in X$}{
        \If{$\ip{w}{\log_{p_0}(x)}\geq 0$}{
            Add $x$ to $S_1$;\;
        }
        \If{$\ip{w}{\log_{p_0}(x)}\leq 0$}{
            Add $x$ to $S_2$;\;
        }
        }
  $S_L$ = FindHull($S_1$,A,B), $S_R$ = FindHull($S_2$,B,A);\;
  \Output {Union($S_L$,$S_R$).}
\end{algorithm2e}

\begin{algorithm2e}
\caption{FindHull}\label{alg:FindHull}
\SetAlgoLined
\DontPrintSemicolon
\SetKwInput{Input}{Input}
\SetKwInOut{Output}{Output}
  \Input{Set of points $S_k$, points $P$ and $Q$.}
  \If{$|S_k|$ is $0$}{
    \Output {$\emptyset$}
  }
  Find the furthest point $F$ from $\gamma_{P\rightarrow Q}$.\;
  Partition $S_k$ into three set $S_0,S_1,S_2$: $S_0$ contains points in $\Delta PFQ$; $S_1$ contains points outside of $\gamma_{P\rightarrow F}$; and $S_2$ contains points outside of $\gamma_{F\rightarrow Q}$.\;
  $S_L$ = FindHull($S_1$,P,F), $S_R$ = FindHull($S_2$,F,Q)\;
  \Output {Union(F,$S_L$,$S_R$);}
\end{algorithm2e}

\section*{Proof of Theorem~\ref{thm:hard-margin-svm-convex} and~\ref{thm:soft-margin-svm-convex}}
Let $x_i\in\mathbb{B}^n$ and let $v_i=\log_{p}(x_i)$ be its its logarithmic map value. The distance between the point and the hyperplane defined by $w\in T_p\mathbb{B}^n$ and $p\in\mathbb{B}^n$ can be written as (see also~(\ref{eq:simp_p2H_tp}))
\begin{align}
\label{eq:svm_proof}
    d(x, H_{w,p}) = \sinh ^{-1}\left(\frac{2 \tanh \left(\sigma_{p}\|v_i\|/2\right)|\langle v_i, w\rangle|}{\left(1-\tanh^2 \left(\sigma_{p}\| v_i\|/2\right)\right)\|w\|\sigma_p\|v_i\|/2}\cdot\frac{\sigma_p}{2}\right).
\end{align}
For support vectors, $|\langle v_i, w\rangle|=1$ and $\|v_i\|\geq 1/\|w\|$. Note that $f(x)=\frac{2\tanh(x)}{x(1-\tanh^2(x))}$ is an increasing function in $x$ for $x>0$ and $g(y)=\sinh^{-1}(y)$ is an increasing function in $y$ for $y\in\mathbb{R}$. Thus, the distance in~(\ref{eq:svm_proof}) can be lower bounded by 
\begin{align}
\label{eq:svm_dist_lower_bound}
    d(x_i,H_{w, p}) \geq \sinh ^{-1}\left(\frac{2 \tanh (\sigma_p/2\|w\|)}{1-\tanh^2(\sigma_p/2\|w\|)}\right).
\end{align}
The goal is to maximize the distance in~(\ref{eq:svm_dist_lower_bound}). To this end observe that $h(x)=\frac{2x}{1-x^2}$ is an increasing function in $x$ for $x\in (0,1)$ and $\tanh (\sigma_p/2\|w\|)\in (0,1)$. So maximizing the distance is equivalent to minimizing $\|w\|$ (or $\|w\|^2$), provided $\sigma_p$ is fixed. Thus the Poincar\'e SVM problem can be converted into the convex problem of Theorem~\ref{thm:hard-margin-svm-convex}; the constraints are added to force the hyperplane to correctly classify all points in the hard-margin setting. The formulation in Theorem~\ref{thm:soft-margin-svm-convex} can also be seen as arising from a relaxation of the constraints and consideration of the trade-off between margin values and classification accuracy.
\vspace{-0.1in}
\section*{Detailed experimental setting}
For the first set of experiments, we have the following hyperparameters. For the Poincar\'e perceptron, there are no hyperparameters to choose. For the Poincar\'e second-order perceptron, we adopt the strategy proposed in~\cite{cesa2005second}. That is, instead of tuning the parameter $a$, we set it to $0$ and change the matrix inverse to pseudo-inverse. For the Poincar\'e SVM and the Euclidean SVM, we set $C=1000$ for all data sets. This theoretically forces SVM to have a hard decision boundary. For the hyperboloid SVM, we surprisingly find that choosing $C=1000$ makes the algorithm unstable. Empirically, $C=10$ in general produces better results despite the fact that it still leads to softer decision boundaries and still breaks down when the point dimensions are large. As the hyperboloid SVM works in the hyperboloid model of a hyperbolic space, we map points from the Poincar\'e ball to points in the hyperboloid model as follows. Let $x\in \mathbb{B}^n$ and $z\in \mathbb{L}^n$ be its corresponding point in the hyperboloid model. Then, 
\begin{align}
    & z_0 = \frac{1-\sum_{i=0}^{n-1}x_i^2}{1+\sum_{i=1}^{n}x_i^2},\;z_j = \frac{2x_j}{1+\sum_{i=1}^{n}x_i^2}\;\forall j\in [n].
\end{align}
On the other hand, Olsson's scRNA-seq data contains $319$ points from $8$ classes and we perform a $70\%/30\%$ random split to obtain training (231) and test (88) point sets. CIFAR10 contains $50,000$ training points and $10,000$ testing points from $10$ classes. Fashion-MNIST contains $60,000$ training points and $10,000$ testing points from $10$ classes. Mini-ImageNet contains $8,000$ data points from $20$ classes and we do $70\%/30\%$ random split to obtain training ($5,600$) and test ($2,400$) point sets. For all data sets we choose the trade-off coefficient $C=5$, and use it with all three SVM algorithms to ensure a fair comparison. We also find that in practice the performance of all three algorithms remains stable when $C\in[1,10]$.

\end{document}